\documentclass[preprint,12pt]{elsarticle}
\usepackage{hyperref} 
\usepackage{tcolorbox}
\usepackage{graphicx}  
\usepackage{multirow}
\usepackage{multicol}
\usepackage{xcolor}
\usepackage{mdframed}
\usepackage{xcolor}
\usepackage{textcomp}
\usepackage{stfloats}
\usepackage{url}
\usepackage{verbatim}
\usepackage[normalem]{ulem}
\usepackage[ruled,vlined]{algorithm2e}
\usepackage{adjustbox}
\usepackage{tabularx}
\usepackage{graphicx}
\usepackage{epstopdf} 
\usepackage{amssymb}
\usepackage{amsmath}
\journal{Applied Energy}

\begin{document}

\begin{frontmatter}



\title{Solar Photovoltaic Assessment with Large Language Model}


\author[label1]{Muhao Guo, Yang Weng} %% Author name
% \author[label2]{Erik Blasch} 
\affiliation[label1]{
% \affiliation{
            organization={Department of Electrical, Computer and Energy Engineering, Arizona State University},%Department and Organization
            addressline={Goldwater Center for Science and Engineering, 650 E Tyler Mall}, 
            city={Tempe},
            postcode={85281}, 
            state={AZ},
            country={USA}}

% \affiliation[label2]{organization={Air Force Office of Scientific Research},
%             addressline={875 North Randolph Street, Suite 325},
%             city={Arlington},
%             postcode={22203},
%             state={VA},
%             country={USA}}

%% Abstract
\begin{abstract}
Accurate detection and localization of solar photovoltaic (PV) panels in satellite imagery is essential for optimizing microgrids and active distribution networks (ADNs), which are critical components of renewable energy systems. 
Existing methods lack transparency regarding their underlying algorithms or training datasets, rely on large, high-quality PV training data, and struggle to generalize to new geographic regions or varied environmental conditions without extensive re-training.
These limitations lead to inconsistent detection outcomes, hindering large-scale deployment and data-driven grid optimization.
In this paper, we investigate how large language models (LLMs) can be leveraged to overcome these challenges. Despite their promise, LLMs face several challenges in solar panel detection, including difficulties with multi-step logical processes, inconsistent output formatting, frequent misclassification of visually similar objects (e.g., shadows, parking lots), and low accuracy in complex tasks such as spatial localization and quantification.
To overcome these issues, we propose the PV Assessment with LLMs (PVAL) framework, which incorporates task decomposition for more efficient workflows, output standardization for consistent and scalable formatting, few-shot prompting to enhance classification accuracy, and fine-tuning using curated PV datasets with detailed annotations.
PVAL ensures transparency, scalability, and adaptability across heterogeneous datasets while minimizing computational overhead. By combining open-source accessibility with robust methodologies, PVAL establishes an automated and reproducible pipeline for solar panel detection, paving the way for large-scale renewable energy integration and optimized grid management.
\end{abstract}

% % \includegraphics{grabs}
% \begin{figure}[h]
%     \centering
%     \includegraphics[width=1\linewidth]{framework.pdf}
%     \caption{Workflow for solar panel detection, localization, and counting using LLMs. The framework combines data engineering, prompt engineering, and fine-tuning to improve accuracy, with outputs including predictions with likelihood and confidence values.}
%     \label{fig:framework}
% \end{figure}

% \begin{highlights}
% \item RA novel framework (PVAL) is proposed that leverages large language models (LLMs) for accurate detection, localization, and quantification of solar panels from satellite imagery.
% \item Combines data engineering, prompt engineering (task decomposition, output standardization, few-shot prompting), and fine-tuning to achieve high accuracy across diverse geographic regions and environmental conditions.
% \item Introduces a confidence-driven auto-labeling mechanism that enables scalable dataset annotation and enhances decision-making for renewable energy integration in microgrids and distribution networks.
% \end{highlights}

%% Keywords

\end{frontmatter}

\section{Introduction}
The global transition toward renewable energy has significantly influenced the operation and planning of power systems, particularly in active distribution networks (ADNs) and microgrids \cite{xiao2023distribution}. As a leading source of clean energy, solar photovoltaic (PV) technology has become increasingly important for decentralizing power generation and enhancing grid resilience \cite{yuan2022determining}. To realize these benefits, accurate detection and mapping of solar PV installations in satellite imagery are critical for efficient energy management, demand-response optimization, and the seamless integration of distributed energy resources (DERs) \cite{guo2023graph}. However, the widespread adoption of solar panels has led to an exponential increase in the amount of imagery data, necessitating highly scalable and adaptable approaches for PV identification \textcolor{black}{\cite{Li2024SolarNetPlus,Zech2024ActivePV}}.

Early approaches to solar panel detection from imagery relied on traditional machine learning models, such as Logistic Regression \cite{abdulmawjood2018detection}, Random Forest \cite{malof2016image}, and Support Vector Machines based methods \cite{yuan2021support}, often employing handcrafted features like texture and color. Although these methods were relatively efficient and required fewer training samples, they lacked the capacity to generalize to diverse spatial patterns. More recent research has turned towards deep learning \cite{wu2022spatial, guo2023continuous}, particularly Convolutional Neural Network (CNN)-based models \cite{ishak2024automated}, which have demonstrated superior performance in extracting vision features from imagery data; \textcolor{black}{recent advances span robust PV module segmentation and city-scale classification
%\cite{Garcia2024Mask2FormerPV,Pereira2024AutomatedPV,Liu2024PKGPVN,Frimane2023SmallPV}.} 
\cite{Alkhatib2025AttentionUNetPV,Ronchetti2025PVPlantsMapping,Luimstra2025ColorElevationPV,Hamzaoglu2025RooftopPV}.}
Notable efforts such as Deep Solar \cite{yu2018deepsolar} showcased large-scale solar panel mapping using Inception-v3, while U-Net \cite{bouazizintegrated} enabled global solar mapping. However, these methods often require substantial annotated datasets \cite{weng2022transform, cui2023sig2vec}\textcolor{black}{—a challenge amplified by roof superstructures and geographic heterogeneity \cite{Li2024SolarNetPlus}—and thus motivate label-efficient strategies \cite{Qiu2024FSRSI,Dong2024FSScene}}, intensive computational resources, and frequent retraining to adapt to changing conditions in different geographical regions and environmental contexts.

These limitations highlight a pressing need for flexible, transparent, and resource-efficient solutions. Large language models (LLMs) have emerged as a promising paradigm for detection tasks with limited labels \cite{luo2022solar}, owing to their extensive pre-training on diverse data sources. Although originally developed for natural language processing (NLP), advanced LLMs such as OpenAI's GPT series \cite{achiam2023gpt}, Google's PaLM \cite{driess2023palm}, and Meta's LLaMA \cite{girdhar2023imagebind} have extended their capabilities to multimodal tasks, including image analysis. 
\textcolor{black}{In remote sensing, vision–language models (VLMs) and datasets are rapidly maturing—e.g., RemoteCLIP\cite{RemoteCLIP2023}, RS5M\cite{RS5M2023}, and RSGPT \cite{RSGPT2023}—alongside broad evaluations and surveys of multimodal LLMs \cite{Huang2024MLLMsurvey,Yin2024MLLMsurvey}.} 
LLMs can capture rich contextual information \cite{guo2024bayesian, guo2023msq, guo2024transparent} and handle heterogeneous inputs and varying contexts, making them particularly suitable for the dynamic environments typical of ADNs and microgrids \cite{xiao2024privacy}.

Although LLMs show great promise for solar panel detection, they face several challenges. 
% \sout{Firstly,} 
LLMs often struggle with multi-step logical processes essential for accurately identifying and assessing solar panels \cite{patel2024multi}, \textcolor{black}{including top-view spatial reasoning required by overhead imagery \cite{Li2024TopViewSpatial}.} 
% \sout{This difficulty arises because solar panel detection typically requires sequential reasoning, such as distinguishing panels from similar structures, understanding spatial relationships, and integrating contextual information from satellite imagery.} 
\textcolor{black}{These challenges stem from sequential disambiguation of confounders, spatial reasoning, and context integration in overhead scenes.} Additionally, LLMs can exhibit inconsistencies in output formatting \cite{stureborg2024large}, which can lead to increased manual intervention and reduced scalability of solar energy management workflows.

Another challenge is the misclassification of visually similar objects. For example, shadows, parking lots, or other flat surfaces may be incorrectly identified as solar panels due to their comparable appearance in satellite images. This issue is exacerbated in environments with high visual complexity or varying lighting conditions, which can obscure or distort the features LLMs rely on for accurate classification\textcolor{black}{; related remote-sensing studies underscore the difficulty of robust shadow handling \cite{ShadowSurvey2024}.} 
% \sout{Furthermore, LLMs often display low accuracy in complex tasks such as spatial localization and quantification of solar panels. Accurate spatial localization requires precise mapping of panel locations, while quantification involves estimating the number and capacity of panels, both of which demand high precision and reliability.} 
\textcolor{black}{These failure modes motivate the task decomposition, schema-standardization, and few-shot strategies introduced below.}

To address the challenges associated with LLMs in solar panel detection, we propose prompting and fine-tuning strategies.
% \sout{First, solar panel detection involves analyzing complex rooftop images to accurately identify and assess solar installations.} 
\textcolor{black}{We operationalize these strategies within an end-to-end PVAL pipeline.} To manage this complexity, we implement task decomposition within our PVAL framework. We begin by defining the overall objective in the prompt, guiding the model to understand the context and goals of the analysis. The task is then broken down into specific steps: first, the model performs a holistic image analysis to detect potential solar panels; second, it pinpoints the exact locations of these panels within predefined regions; and third, it estimates the quantity of panels based on their arrangement and visibility. This structured approach ensures that each sub-task is handled systematically, reducing the likelihood of errors and improving the overall accuracy of detection. 
% \sout{By decomposing the task, we enable the LLM to focus on manageable components, thereby enhancing its ability to process complex visual data effectively.}

Second, consistent and interpretable outputs are crucial for integrating detection results into existing solar energy management systems (EMS). To achieve this, we employ output standardization by defining a JSON-based response format. This standardized structure includes key fields such as \textit{``solar panels present"}, \textit{``location"}, and \textit{``quantity"}, each with predefined possible values to ensure uniformity. For example, the \textit{``location"} field uses a detailed classification schema that divides the image into specific regions like \textit{``top-left"} or \textit{``center"} as illustrated in our visual schema.
% (Figure \ref{fig:10labels})
This standardization addresses the problem of inconsistent output formats, facilitating seamless data integration into geographic information systems (GIS) and energy management platforms. By ensuring that all outputs adhere to a uniform format, we eliminate the need for additional preprocessing, thereby streamlining workflows and enhancing the scalability of our detection system. \textcolor{black}{In practice, this schema reduces format drift and makes outputs directly machine-actionable for downstream GIS/EMS pipelines.}

Third, accurate classification of solar panels, especially when distinguishing them from visually similar objects like shadows or parking lots, is a significant challenge. To address this issue, we utilize few-shot prompting by incorporating a limited number of highly relevant examples within the prompt. These examples include both positive cases (where solar panels are present) and negative cases (where they are absent). 
% \sout{providing the LLM with clear references on how to handle different scenarios.} 
By using five-shot prompting, we enhance the model's ability to generalize from these examples without extensive labeled datasets\textcolor{black}{, consistent with findings in label-efficient remote sensing and multimodal few-shot learning \cite{Qiu2024FSRSI,Dong2024FSScene,DiffCLIP2024}.} \textcolor{black}{We also curate exemplars that mirror common confounders (e.g., shadows, HVAC units) to improve robustness.} This approach mitigates misclassification issues, enabling the LLM to better distinguish between similar-looking objects and thereby increasing the reliability of solar panel detection.

% \sout{Last, another challenge for LLMs in solar panel detection is handling complex spatial localization and quantification tasks.} 
Accurate spatial localization requires precise mapping of panel locations, while quantification involves estimating the number and capacity of panels, both of which demand high precision and reliability. To overcome these limitations, we investigate fine-tuning strategies within our PVAL framework that complement the prompting techniques. By fine-tuning the LLM on specialized datasets that include annotated satellite images with detailed spatial information, the model can better understand and perform multi-step reasoning required for localization and quantification. \textcolor{black}{Fine-tuning complements prompting by aligning the model with rooftop-specific patterns and spatial cues.}

An additional benefit of PVAL is its capacity to provide nuanced likelihood and confidence values for identified solar panels. While conventional computer vision techniques have explored confidence scoring \cite{blasch2008image}, LLMs can produce richer, context-aware assessments \cite{xu2024sayself}. These confidence and likelihood metrics enable auto-labeling of large-scale datasets, reducing the need for manual annotation and enhancing system scalability\textcolor{black}{; they also connect to contemporary work on calibration and uncertainty estimation in deep learning and NLP \cite{Wang2023Calibration,Geng2024NLPCalibration,UQSurvey2024}.} \textcolor{black}{Operationally, we use these scores to gate auto-labeled samples and to prioritize uncertain tiles for targeted review.} Consequently, powergrid operators can efficiently manage extensive data, ensuring accurate and reliable integration into ADNs and microgrids.

In this work, we explore solar panel detection by proposing PVAL, which combines large language models (LLMs) with data engineering, prompt engineering, and fine-tuning strategies. The key contributions of this paper are:
%%%%%%%%%%

\begin{enumerate}
\item \textbf{Prompting Strategies for PV Detection}: We address challenges such as misclassification and inconsistent outputs by implementing a combination of task decomposition, standardized output formatting, and few-shot prompting. These strategies enhance the interpretability of results and optimize the model’s ability to understand spatial and numerical information in satellite imagery, thereby improving detection accuracy and reliability.
\item \textbf{Robust Fine-tuning Techniques}: To mitigate geographic and environmental variability, we employ robust fine-tuning methods within our PVAL framework. By fine-tuning the LLM on specialized, curated datasets that include annotated satellite images with detailed spatial information, we ensure consistent performance across diverse regions and varying imaging conditions.
\item \textbf{Scalable and Adaptable Framework}: PVAL scales seamlessly across diverse U.S. regions, achieving high accuracy not only in detecting solar panels but also in localizing and quantifying them. This adaptability ensures robust performance in varied spatial and regional contexts, making it ideal for large-scale applications.
\item \textbf{Confidence-Driven Auto-Labeling Mechanism}: We introduce an auto-labeling mechanism driven by confidence and likelihood metrics. This mechanism enables the automatic labeling of large-scale datasets, significantly reducing the need for manual annotation. Consequently, powergrid operators can efficiently manage extensive data, ensuring accurate and reliable integration into Advanced Distribution Networks (ADNs) and microgrids.
\end{enumerate}

Harnessing LLMs’ multimodal capabilities to detect solar panels and bridging prompt engineering and fine-tuning techniques, this research lays the groundwork for more intelligent, adaptive, and scalable solutions. We aim to facilitate the sustainable transformation of distribution networks and microgrids, supporting broader integration of renewable energy sources and paving the way towards smarter, more resilient power systems.

\section{Problem Description}
The widespread adoption of photovoltaic (PV) systems necessitates scalable, accurate, and efficient methodologies for their detection, localization, and quantification using satellite imagery. 
Current detection approaches encounter challenges due to the vast spatial heterogeneity in rooftop structures, varying environmental conditions, and the inherent limitations of traditional detection systems. Traditional convolutional neural network (CNN)-based models, while effective in feature extraction, require extensive annotated datasets and frequent retraining to adapt to different geographical regions and environmental contexts. Moreover, these models often struggle with the misclassification of visually similar objects—such as shadows, parking lots, and flat surfaces—which leads to reduced accuracy and increased manual intervention.

Formally, let the input satellite image be represented as \( I \). The objective is to develop a function \( f(I) \) that maps the input image to a set of outputs \( \{ \textit{D}, \textit{L}, \textit{Q}, \textit{C} \} \), where \( \textit{D} \) indicates the binary detection of the presence of solar panels, \( \textit{L} \) specifies the spatial localization of panels within the image \( I \), \( \textit{Q} \) represents the quantification of the number of solar panels in the image, and \( \textit{C} \) provides a credibility score representing the likelihood of correct detection, where \( C \in [0, 1] \). The solution must ensure robust performance across diverse geographical regions, rooftop configurations, and varying environmental conditions. Additionally, the outputs \( \{ \textit{D}, \textit{L}, \textit{Q}, \textit{C} \} \) must be interpretable and seamlessly integrable into decision-making frameworks for active distribution networks (ADNs) and microgrids.

This work investigates how to leverage LLMs to process satellite imagery for PV detection. By incorporating data engineering, prompt engineering, and fine-tuning, the proposed PVAL efficiently detects, localizes, and quantifies solar panels while generating structured outputs with confidence metrics. PVAL emphasizes scalability, adaptability, and minimal reliance on manual annotations, providing a transformative solution for grid planning, renewable energy integration, and operational optimization.

\section{Methodology}
\label{sec:methodology}
\begin{figure}[h]
    \centering
    \includegraphics[width=1\linewidth]{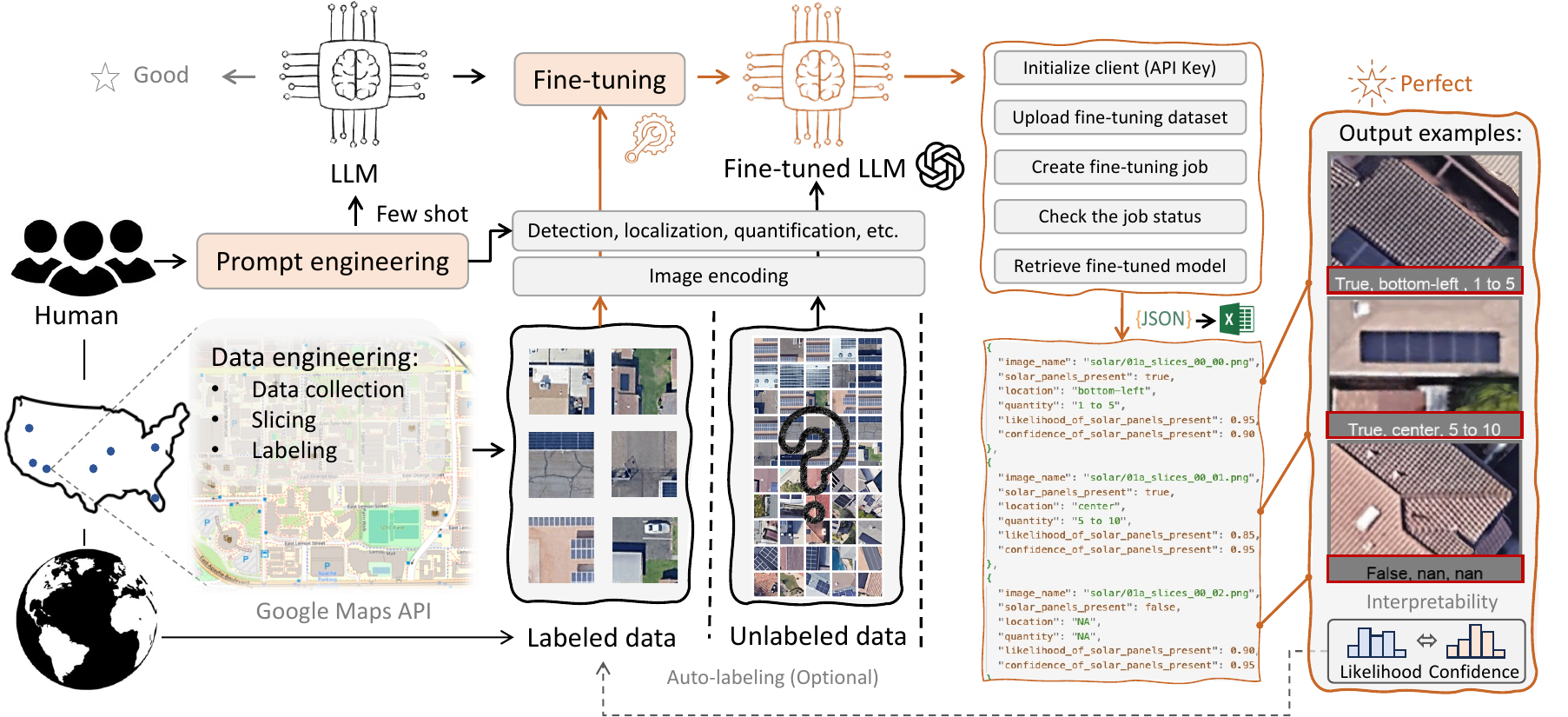}
    \caption{\textcolor{black}{Workflow of PVAL: solar panel detection, localization, and counting using LLMs.} The framework combines data engineering, prompt engineering, and fine-tuning to improve accuracy, with outputs including predictions with likelihood and confidence values.}
    \label{fig:framework}
\end{figure}

This section describes how PVAL can leverage LLMs for solar panel detection. PVAL incorporates data engineering, prompt engineering strategies, fine-tuning, and a dual-metric auto-labeling mechanism. Figure~\ref{fig:framework} illustrates the complete workflow of our approach.

\subsection{Data Engineering}
Data engineering is a foundational component of PVAL ensuring that the model is trained and evaluated on high-quality, well-structured datasets. PVAL incorporates three stages: data collection, image slicing, and labeling.

\subsubsection{Data Collection}
We obtain geographic coordinates of solar panel installations from OpenStreetMap (OSM) using the Overpass API.{\color{black}\footnote{https://overpass-api.de/}} 
{\color{black}For each region, we preserve the full Overpass query, server endpoint, and export timestamp, which together define the exact OSM snapshot used.}
These coordinates serve as anchors for retrieving satellite imagery from the Google Maps Static API{\color{black}\footnote{https://developers.google.com/maps/documentation/maps-static/overview}}{\color{black}, with parameters fixed at \texttt{maptype=satellite}, \texttt{zoom=20}, and \texttt{size=400x400}.}
{\color{black}Google imagery is publicly accessible with an API key, but redistribution is restricted by licensing.}
{\color{black}To ensure reproducibility, we release all metadata—coordinates, queries, request parameters, timestamps, and per-image hashes—so that the dataset can be regenerated by any user with their own API key.}
We target diverse U.S. regions (Seattle, WA; Orlando, FL; Osage Beach, MO; Harlem, NY; Tempe, AZ; and Santa Ana, CA) to capture a wide spectrum of climatic conditions, urban densities, and architectural styles, thereby enhancing the model’s generalization ability.

% \subsubsection{Data Collection}
% We obtain geographic coordinates of solar panel installations from \textbf{OpenStreetMap (OSM)} using the Overpass API.\footnote{https://overpass-api.de/} 
% For each region, we preserve the full Overpass query, server endpoint, and export timestamp, which together define the exact OSM snapshot used. 
% These coordinates serve as anchors for retrieving satellite imagery from the \textbf{Google Maps Static API},\footnote{https://developers.google.com/maps/documentation/maps-static/overview} with parameters fixed at \texttt{maptype=satellite}, \texttt{zoom=20}, and \texttt{size=400x400}. 
% Google imagery is publicly accessible with an API key, but redistribution is restricted by licensing. 
% To ensure reproducibility, we release all metadata—coordinates, queries, request parameters, timestamps, and per-image hashes—so that the dataset can be regenerated by any user with their own API key. 
% We target diverse U.S. regions (Seattle, WA; Orlando, FL; Osage Beach, MO; Harlem, NY; Tempe, AZ; and Santa Ana, CA) to capture a wide spectrum of climatic conditions, urban densities, and architectural styles, thereby enhancing the model’s generalization ability.

\subsubsection{Image Slicing}
{\color{black}
Each high-resolution image is divided into a fixed $4 \times 4$ grid, yielding 16 tiles per rooftop scene. 
This systematic slicing not only expands the training corpus but also preserves resolution at a finer spatial scale. 
Small-scale photovoltaic arrays remain visible, enabling the model to detect presence (D), localize placement (L), and estimate quantity (Q). 
The slicing process is deterministic, and we release both the slicing script and the metadata schema so that identical tile indices can be reconstructed.
}

% \subsubsection{Data Collection}
% The process begins by identifying geographic coordinates of solar panel installations from OpenStreetMap (OSM) via the Overpass API.
% % ~\cite{overpass_api}. 
% These coordinates guide the retrieval of high-resolution satellite imagery using the Google Maps Static API.
% % ~\cite{google_maps_static_api}. 
% To ensure representativeness, we target diverse U.S. regions: Seattle, WA; Orlando, FL; Osage Beach, MO; Harlem, NY; Tempe, AZ; and Santa Ana, CA. Capturing a wide array of climatic conditions, urban densities, and architectural styles enhances the model’s ability to generalize across heterogeneous environments (see Figure~\ref{fig:US_map}).

% \subsubsection{Image Slicing}
% To enhance detection granularity and scalability, we divide each high-resolution satellite image into a $4 \times 4$ grid, resulting in 16 equally sized tiles. The systematic slicing increases the number of training examples while preserving image resolution. Smaller solar panels remain distinguishable, enabling the model to identify, localize, and count panels at a finer spatial scale.

\subsubsection{Data Labeling}
We employ manual annotation to label each tiled image, recording the presence, location, and estimated quantity of solar panels. 
% {\color{black}We clarify that four annotators participated in the labeling. For every image, at least two annotators labeled independently, and we observed 100\% pairwise agreement among all annotators, corresponding to a Kappa score of $1.0$.}
{\color{black}Four annotators participated in the labeling. All annotators had prior experience in rooftop PV identification and visual interpretation of overhead imagery, and they received dedicated training before the annotation process. For every image, at least two annotators labeled independently, and we observed 100\% pairwise agreement among all annotators, corresponding to a Kappa score of $1.0$.}
The human-in-the-loop approach ensures high-quality ground truth labels, facilitating effective fine-tuning and reliable performance evaluations \cite{blasch2012high}.

\subsection{Model Architecture and Input Encoding}
The proposed PVAL system leverages the capabilities of GPT-4o, a large multimodal model built upon the transformer architecture, renowned for its ability to process and reason over both textual and visual inputs through integrated attention mechanisms \cite{vaswani2017attention}. This section outlines the core model components and how satellite imagery is encoded for solar panel detection.

GPT-4o extends traditional transformer architectures by incorporating a vision encoder that transforms raw images into high-level embeddings. These visual embeddings are then integrated with language tokens, enabling unified multimodal reasoning. Unlike conventional vision transformers that apply self-attention directly to patches of image data, GPT-4o relies on a pretrained image encoder to process satellite imagery, followed by transformer layers that jointly reason over visual and textual information.

For the text modality, the core transformer uses self-attention mechanisms, formulated as:
\begin{equation}
\text{Attention}(Q, K, V) = \text{Softmax}\left(\frac{QK^T}{\sqrt{d_k}}\right)V,
\end{equation}
with query, key, and value matrices \( Q, K, V \) derived from token embeddings. Multi-head attention extends this formulation as:
\begin{equation}
\text{MHA}(Q, K, V) = \text{Concat}(\text{head}_1, \ldots, \text{head}_h)W^O,
\end{equation}
allowing the model to attend to different subspaces in parallel.

Position-wise feed-forward networks (FFNs) and residual connections, combined with layer normalization, enhance non-linear representation capacity:
\begin{equation}
\text{FFN}(x) = \text{ReLU}(xW_1 + b_1)W_2 + b_2.
\end{equation}

For input processing, satellite images are uploaded and internally encoded by GPT-4o's vision subsystem. Although the raw image data is transmitted using formats such as \textit{base64 encoding} \cite{josefsson2006base16}, the model does not process base64 strings directly. Instead, images are converted into embeddings that the model can use in conjunction with textual prompts.

The prompts are carefully structured to guide the model’s attention toward image understanding tasks such as detecting, localizing, and quantifying solar panels. Outputs are returned as structured JSON, capturing information such as panel presence, approximate count, and bounding box location. This format facilitates integration with renewable energy monitoring platforms.

By combining prompt engineering with multimodal attention, PVAL enables GPT-4o to generalize across varied environmental and geographic conditions, achieving reliable solar panel detection from satellite imagery.

\subsection{Prompting Strategies for PV Detection}
\begin{figure}[t]
    \centering
    \includegraphics[width=1 \linewidth]{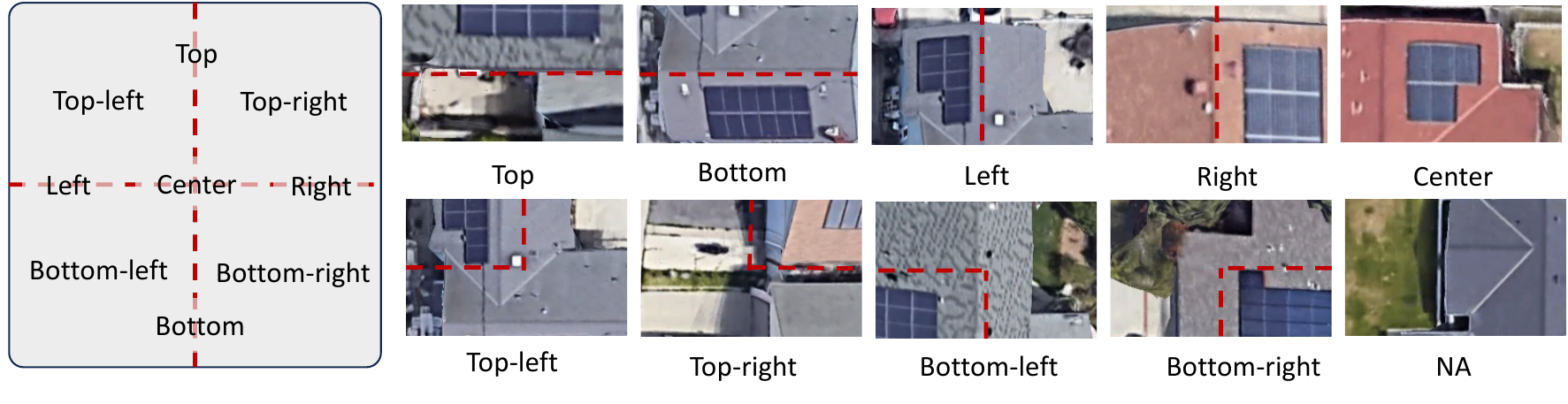}
    \caption{
    Illustration of all possible solar panel locations as defined in the work, accompanied by representative image examples for each category. The locations include \textit{Top}, \textit{Bottom}, \textit{Left}, \textit{Right}, \textit{Center}, \textit{Top-left}, \textit{Top-right}, \textit{Bottom-left}, \textit{Bottom-right}, and \textit{``NA"} (indicating no solar panels). 
    }
    \label{fig:10labels}
\end{figure} 
Accuracy, scalability, minimizing ambiguity, and interpretability are critical requirements for applying LLMs to solar panel detection in satellite imagery. Accuracy ensures precise identification and classification of solar panels, while scalability allows the approach to handle large and diverse datasets efficiently. Minimizing ambiguity in outputs is crucial for reliable interpretation, and interpretability ensures that the generated results align seamlessly with downstream applications such as geospatial mapping and performance evaluation.

We address these requirements through prompt engineering strategies incorporating task decomposition \cite{khot2022decomposed}, 
% li2024drattack
output standardization \cite{chu2024better}, and few-shot prompting \cite{brown2020language}.
% schick2020exploiting
Task decomposition allows the model to break down complex tasks into manageable sub-tasks, guiding the LLM to systematically identify solar panel presence, location, and quantity. Output standardization ensures consistent and interpretable results through a predefined JSON structure, minimizing variability and ambiguity in the outputs. Few-shot prompting enhances the model’s adaptability by providing illustrative examples, allowing it to generalize across diverse scenarios without extensive retraining.

\subsubsection{Task Decomposition}
\begin{tcolorbox}[
    colback=gray!5!white,
    colframe=gray!75!black,
    title=\scriptsize\bfseries Prompt -- Task Decomposition,
    fonttitle=\bfseries,
    width=1\linewidth,
    boxrule=0.5mm,% Thickness of the border
    left=1.5 mm,   % Space between the left border and text
    right=0mm,    % Space between the right border and text
    top=0mm,      % Space between the top border and text
    bottom=-2mm,   % Space between the bottom border and text
]
\scriptsize\ttfamily
\label{prompt:Task Description and Steps}
Identify the \colorbox{red!10}{presence} of solar panels in images of residential rooftops, and determine their \colorbox{orange!10}{locations} and \colorbox{yellow!10}{quantity} \colorbox{gray!20}{within the images}. 
You will be provided with images that may contain residential rooftop solar systems. Analyze each image to detect solar panels.
\\
\colorbox{gray!20}{\textbf{Steps:}} \\
1.\colorbox{red!10}{**Image Analysis**}: Examine the entire image to identify any objects that appear to be solar panels.\\
2.\colorbox{orange!10}{**Panel Location**}: Determine the coordinates or area within the image where the solar panels are located.\\
3.\colorbox{yellow!10}{**Panel Quantification**}: Calculate or estimate the number of solar panels based on their appearance and arrangement.\\
% \normalsize\normalfont
% \footnotesize
\scriptsize
\end{tcolorbox}
The prompt begins with a clear task description, outlining the objective of analyzing rooftop images to detect solar panels, as shown in the Prompt box \ref{prompt:Task Description and Steps}. The description ensures the model understands the task's overall goal and context. The process is detailed through steps that guide the model to examine image holistically, identify solar panel locations using predefined regions, and estimate the PV quantity based on arrangement and visibility. The progression breaks the task into manageable sub-tasks, enabling a structured approach.

\subsubsection{Output Standardization}
\label{sec:Output Standardization}
\begin{tcolorbox}[
    colback=gray!5!white,
    colframe=gray!75!black,
    title=\scriptsize\bfseries Prompt -- Output Standardization,
    fonttitle=\bfseries,
    width=1\linewidth,
    boxrule=0.5mm,% Thickness of the border
    left=1.5mm,   % Space between the left border and text
    right=0mm,    % Space between the right border and text
    top=0mm,      % Space between the top border and text
    bottom=-2mm,   % Space between the bottom border and text
]
\label{Prompt -- Output Standardization}
\scriptsize\ttfamily
The output should be in \colorbox{gray!20}{JSON format}, structured as follows, with each field restricted to specific \colorbox{gray!20}{possible values} for consistency and accuracy: \\
\colorbox{red!10}{"solar\_panels\_present"}: A boolean value indicating if solar panels are detected.
Possible values: \colorbox{gray!20}{[true, false]} \\
\colorbox{orange!10}{"location"}: A description or coordinates indicating where the panels are located within the image.
Possible values: \colorbox{gray!20}{[left, right, bottom, top, top-left, top-right} \colorbox{gray!20}{bottom-right, bottom-left, center, NA]} \\
\colorbox{yellow!10}{"quantity"}: The number of solar panels detected in the image.
Possible values: \colorbox{gray!20}{[0 to 1, 1 to 5, 5 to 10, 10 to inf, NA]} \\
\colorbox{blue!10}{"likelihood\_of\_solar\_panels\_present"}: A value indicating the probability of solar panels being present.
Possible values: A decimal range \colorbox{gray!20}{from 0.00 to 1.00}\\
\colorbox{green!10}{"confidence\_of\_solar\_panels\_present"}: A value indicating the model's confidence in its prediction.
Possible values: A decimal range \colorbox{gray!20}{from 0.00 to 1.00}\\
% \footnotesize
\scriptsize	
\end{tcolorbox}
To ensure consistent and interpretable outputs, we define a standardized JSON-based response format. The JSON structure is shown in Prompt box \ref{Prompt -- Output Standardization} including essential fields such as \textit{``solar panels present"}, which indicates whether solar panels are detected; \textit{``location"}, which specifies the spatial region or quadrant where the panels are located; and \textit{``quantity"}, which estimates the number of panels. In addition, fields of \textit{``likelihood of solar panels present"} and \textit{``confidence of solar panels present"} provide quantitative measures of the reliability of the predictions, enhancing interpretability and enabling users to assess the strength and uncertainty of the outputs \cite{jousselme2023uncertain}.

The design of the \textit{``location"} field incorporates a detailed classification schema for spatial localization, dividing the image into regions such as \textit{``Top"}, \textit{``Bottom"}, \textit{``Left"}, \textit{``Right"}, \textit{``Center"}, and quadrants like \textit{``Top-left"} and \textit{``Bottom-right"}, etc. The spatial schema is illustrated in Figure \ref{fig:10labels}, which provides a visual representation of all possible locations with example images for each category. These examples showcase common rooftop configurations where solar panels may be present or absent, such as panels centralized on a roof or distributed across specific quadrants. For cases where no solar panels are detected, the \textit{``location"} field is assigned the value \textit{``NA"}, ensuring completeness and consistency in the output.

The PVAL classification schema plays a critical role in the design of the prompt, as it enables the LLM to focus on specific spatial patterns and ensures that the predictions are both interpretable and actionable. By dividing the image into discrete regions and associating these regions with representative examples, the model gains a structured understanding of spatial relationships. Dividing the image not only improves the clarity of the outputs but also enhances the model's ability to generalize across diverse rooftop configurations and environmental conditions. The structured prompt, coupled with the spatial schema, establishes a clear connection between the input imagery and the model's output, ensuring that the predictions align with real-world scenarios. In addition, we set the “temperature” parameter for LLM to zero to ensure consistency.

\subsubsection{Few-shot Prompting}
\begin{tcolorbox}[
    colback=gray!5!white,
    colframe=gray!75!black,
    title=\scriptsize\bfseries Prompt -- Few-shot,
    fonttitle=\bfseries,
    width=1\linewidth,
    boxrule=0.5mm,% Thickness of the border
    left=1.5mm,   % Space between the left border and text
    right=0mm,    % Space between the right border and text
    top=0mm,      % Space between the top border and text
    bottom=-2mm,   % Space between the bottom border and text
]
\label{Prompt -- Few-shot}
\scriptsize\ttfamily % Adjust font size between tiny and small
\colorbox{gray!20}{$\#$ Example 1 (Solar)}: \\
\{ \colorbox{red!10}{"solar\_panels\_present"}: true, \\
\mbox{} \mbox{} \colorbox{orange!10}{"location"}: "top-left", \\
\mbox{} \mbox{} \colorbox{yellow!10}{"quantity"}: "0 to 1", \\
\mbox{} \mbox{} \colorbox{blue!10}{"likelihood\_of\_solar\_panels\_present"}: 0.98, \\
\mbox{} \mbox{}  \colorbox{green!10}{"confidence\_of\_solar\_panels\_present"}: 0.90 \} \\
% \colorbox{gray!20}{$\#$ Example 2}: \\
% \{ \colorbox{red!10}{"solar\_panels\_present"}: true, \\
% \mbox{} \mbox{} \colorbox{orange!10}{"location"}: "center", \\
% \mbox{} \mbox{}  \colorbox{yellow!10}{"quantity"}: "1 to 5", \\
% \mbox{} \mbox{}  \colorbox{blue!10}{"likelihood\_of\_solar\_panels\_present"}: 0.78, \\
% \mbox{} \mbox{}  \colorbox{green!10}{"confidence\_of\_solar\_panels\_present"}: 0.65 \} \\
\colorbox{gray!20}{$\#$ Example 2 (No Solar)}: \\
\{ \colorbox{red!10}{"solar\_panels\_present"}: false, \\
\mbox{} \mbox{}  \colorbox{orange!10}{"location"}: "NA", \\
\mbox{} \mbox{}  \colorbox{yellow!10}{"quantity"}: "NA", \\
\mbox{} \mbox{} \colorbox{blue!10}{"likelihood\_of\_solar\_panels\_present"}: 0.21, \\
\mbox{} \mbox{}  \colorbox{green!10}{"confidence\_of\_solar\_panels\_present"}: 0.87 \} \\
% \normalsize\normalfont
% \colorbox{gray!20}{$\#$ ... ...}
% \footnotesize
\scriptsize	
\end{tcolorbox}
To improve generalization and adaptability, we integrate few-shot examples within the prompt. As shown in Prompt box \ref{Prompt -- Few-shot}, these examples demonstrate both positive and negative cases, helping the model generalize across a variety of scenarios. For instance, one example shows the case of solar panels being detected, with their location specified as the \textit{``top-left"} and their quantity as a range, such as \textit{``0 to 1"}. Another example illustrates the absence of solar panels, with fields like \textit{``location"} and \textit{``quantity"} marked as \textit{``NA"}. These examples serve as a reference for the LLM, clarifying how to handle different inputs and cases. In practice, we use five-shot prompting. 

The prompt engineering strategies in this work establish a scalable framework for applying LLMs to complex geospatial tasks. By integrating task decomposition, output standardization, and few-shot prompting, PVAL aligns LLM capabilities with the demands of large-scale solar panel detection.

\begin{tcolorbox}[
    colback=gray!5!white,
    colframe=gray!75!black,
    title=\scriptsize\bfseries A complete prompt,
    fonttitle=\bfseries,
    width=1\linewidth,
    boxrule=0.5mm,
    left=1.5mm,
    right=0mm,
    top=0mm,
    bottom=-2mm
]
\scriptsize\ttfamily
\color{black}
% ===== Task Decomposition (verbatim content with colors removed) =====
Identify the presence of solar panels in images of residential rooftops, and determine their locations and quantity within the images. 
You will be provided with images that may contain residential rooftop solar systems. Analyze each image to detect solar panels.
\\
\textbf{Steps:} \\
1. **Image Analysis**: Examine the entire image to identify any objects that appear to be solar panels.\\
2. **Panel Location**: Determine the coordinates or area within the image where the solar panels are located.\\
3. **Panel Quantification**: Calculate or estimate the number of solar panels based on their appearance and arrangement.\\[1pt]

% ===== Output Standardization (verbatim content with colors removed) =====
% \textbf{Output Standardization}\\
The output should be in JSON format, structured as follows, with each field restricted to specific possible values for consistency and accuracy: \\
"solar\_panels\_present": A boolean value indicating if solar panels are detected. Possible values: [true, false] \\
"location": A description or coordinates indicating where the panels are located within the image. Possible values: [left, right, bottom, top, top-left, top-right, bottom-right, bottom-left, center, NA] \\
"quantity": The number of solar panels detected in the image. Possible values: [0 to 1, 1 to 5, 5 to 10, 10 to inf, NA] \\
"likelihood\_of\_solar\_panels\_present": A value indicating the probability of solar panels being present. Possible values: a decimal range from 0.00 to 1.00 \\
"confidence\_of\_solar\_panels\_present": A value indicating the model's confidence in its prediction. Possible values: a decimal range from 0.00 to 1.00 \\
% --- Added likelihood/confidence supplemental definitions (as requested) ---
"likelihood\_of\_solar\_panels\_present": A quantitative measure of how probable it is that a certain hypothesis or model predicts the observed data. The value should be a decimal range from 0.0 to 1.0. 0.0 means a low likelihood, and 1 means a high likelihood. \\
Possible values for the likelihood of solar\_panels\_present are from 0 to 1. \\
"confidence\_of\_solar\_panels\_present": The strength of the model's belief in its prediction. The value should be a decimal range from 0.0 to 1.0. 0.0 means a low confidence for the answer, and 1 means a high confidence for the answer. \\
Possible values for the confidence of solar\_panels\_present are from 0 to 1. \\[1pt]

The output should have the following style:\\
\# Example 1 (Solar): \\
\{ "solar\_panels\_present": true, \\
\mbox{} \mbox{} "location": "top-left", \\
\mbox{} \mbox{} "quantity": "0 to 1", \\
\mbox{} \mbox{} "likelihood\_of\_solar\_panels\_present": 0.98, \\
\mbox{} \mbox{} "confidence\_of\_solar\_panels\_present": 0.90 \} \\
\# Example 2 (No Solar): \\
\{ "solar\_panels\_present": false, \\
\mbox{} \mbox{} "location": "NA", \\
\mbox{} \mbox{} "quantity": "NA", \\
\mbox{} \mbox{} "likelihood\_of\_solar\_panels\_present": 0.21, \\
\mbox{} \mbox{} "confidence\_of\_solar\_panels\_present": 0.87 \} \\
\scriptsize
\end{tcolorbox}

\subsection{Robust Fine-Tuning Techniques}
Adapting LLMs to specialized tasks like solar panel detection in satellite imagery requires domain-specific fine-tuning \cite{tinn2023fine}. 
% tinn2023fine ding2023parameter jeong2024fine
Although LLM excels at general-purpose understanding, it has to be refined to interpret complex rooftops, diverse imaging conditions, and varied environmental factors.
% Fine-tuning enables the model to generate structured outputs: presence (\textit{``True"} or \textit{``False"}), spatial location (e.g., \textit{``top-left"}), and quantity (e.g., \textit{``1 to 5"}), which are critical for downstream renewable energy analytics.

We employed OpenAI’s API infrastructure for the fine-tuning process, preparing the dataset in JSON Lines (JSONL) format. Each entry contained base64-encoded images, the proposed prompts, and ground truth labels, ensuring a consistent and reproducible pipeline. For the classification task, the supervised fine-tuning procedure optimized model parameters via cross-entropy loss, penalizing misclassifications and reinforcing accurate predictions:
\begin{equation}
\mathcal{L} = -\frac{1}{N} \sum_{i=1}^{N} \left[y_i \log(\hat{y}_i) + (1 - y_i) \log(1 - \hat{y}_i)\right],
\end{equation}
where $N$ is the total number of samples, $y_i$ represents the ground truth label for the $i$-th sample, $\hat{y}_i$ denotes the predicted probability of the positive class for the $i$-th sample.
The cloud-based workflow is efficiently scaled to large, heterogeneous datasets, making it easily adaptable to other geospatial analysis tasks. 
% Although confidence scoring occurs at inference time, the fine-tuning step lays the groundwork for producing not only accurate but also interpretable results. 
By bridging the gap between general LLM capabilities and specific geospatial requirements, the fine-tuned GPT-4o model underscores the promise of LLMs in large-scale solar panel detection and energy resource planning.

\subsection{Likelihood and Confidence Mechanism for Auto-Labeling}
% To enhance the understanding of prediction reliability, PVAL's output includes both likelihood and confidence metrics. The metrics provide complementary insights.

% PVAL harnesses both likelihood and confidence metrics to automatically label large-scale datasets, thereby reducing manual annotation overhead.
PVAL harnesses both likelihood and confidence metrics to automatically label large-scale datasets, that can potentially be utilized for semi-supervised learning, thereby reducing manual annotation overhead.

Likelihood quantifies the probability that solar panels are present in the imagery. Higher likelihood values indicate stronger evidence supporting the presence of solar panels. Specifically, values approaching $1$ suggest a strong match to features characteristic of solar panels, implying a high probability of their presence. Conversely, values closer to $0$ denote a weak or negligible match, suggesting their absence.

Confidence reflects the language model’s internal certainty regarding its identification. Confidence values near $1$ signify that the model is highly assured in its prediction, while values closer to $0$ indicate greater uncertainty about the decision.

Likelihood and confidence metrics provide complementary insights for the auto-labeling process. In a true positive case, high likelihood combined with high confidence indicates that the model both detects solar panels accurately and does so with strong conviction, reinforcing the reliability of the automatic label. In contrast, a false negative with low confidence may highlight challenging conditions—such as complex rooftop geometries or poor image quality—thereby prompting operators to verify the label through additional data or manual inspection.

By employing this dual-metric approach in the auto-labeling mechanism, power grid operators can manage extensive datasets, ensuring accurate and reliable information integration into ADNs and microgrids.
High-confidence predictions can be utilized for automated labeling, streamlining the process of annotating data for future model training. In contrast, low-confidence predictions can be flagged for manual review, enabling further data collection or targeted model refinement to enhance overall system performance.
This flexible, transparent framework empowers stakeholders to make data-driven decisions in monitoring and managing distributed renewable energy resources.

{\color{black} To summarize the overall workflow, we provide a step-by-step description of the pipeline with fine-tuning in Algorithm~\ref{algorithm:PVAL}.
}

\begin{algorithm}[t]
\label{algorithm:PVAL}
\DontPrintSemicolon
\SetAlCapFnt{\small}\SetAlCapNameFnt{\small} 
\scriptsize
\color{black}
\caption{PVAL Pipeline with Fine-Tuning Process}
\KwIn{Regions $\mathcal{R}$, Overpass query $\mathcal{Q}$, MapsKey, grid $G{=}4{\times}4$, schema $\mathcal{S}$, few-shot $\mathcal{E}$, temperature $T{=}0$; thresholds $(\tau_p,\tau_c)$, OpenAI \text{api\_key}; base\_model (e.g., \text{gpt-4o}); hyperparams (epochs = 5)}
\KwOut{Fine-tuned model id $\theta^\ast$}
\textbf{1) Collect imagery:} Query Overpass$(\mathcal{R},\mathcal{Q})\!\to$ coords; fetch rooftop images via MapsKey.\\
\textbf{2) Tile \& seed labels:} Slice each image into $G$ tiles; attach human labels $(D,L,Q)$ where available, else $\varnothing$.\\
\textbf{3) Prompted inference (unlabeled tiles):} For each unlabeled tile $t$: build $P=\text{AssemblePrompt}(t,\mathcal{E},\mathcal{S})$; run $\hat{y}=\text{GPT4o}(P,T)$.
\\
\textbf{4) (Optional) confidence auto-labeling:} If 
$\widehat{p}\!\ge\!\tau_p \land \widehat{c}\!\ge\!\tau_c$, fill missing label with $(\widehat{D},\widehat{L},\widehat{Q})$\\
\textbf{5) Build fine-tuning set:} $\mathcal{D}_{\text{ft}}=\{(t,y):y\neq\varnothing\}$; serialize JSONL lines \emph{(base64 image, prompt with $\mathcal{E}$ and $\mathcal{S}$, target JSON $\{D,L,Q\}$)} as \text{solar.jsonl}.\\

\textbf{6) Fine-tuning via OpenAI API:}\\
\Indp
\textit{(a) Initialize client)}: $\text{client} \leftarrow \text{OpenAI(api\_key)}$.\\
\textit{(b) Upload training file}: $\text{file\_id} \leftarrow \text{client.files.create(file=solar.jsonl, purpose="fine-tune").id}$.\\
\textit{(c) Create job}: $\text{job} \leftarrow \text{client.fine\_tuning.jobs.create(file\_id, base\_model, n\_epochs, batch, lr})$.\\
\textit{(d) Poll status)}: \While{\text{status} $\notin$ \{\text{succeeded}, \text{failed}\}}{
  $\text{job} \leftarrow \text{client.fine\_tuning.jobs.retrieve(job.id)}$; \ \text{status} $\leftarrow$ \text{job.status}; \ 
}
\textit{(e) On success)}: $\theta^\ast \leftarrow \text{job.fine\_tuned\_model}$; \text{result\_files} $\leftarrow$ \text{job.result\_files}.\\
\Indm
\textbf{7) Evaluate \& use fine-tuned model:} Evaluate on held-out set at $T{=}0$ (det: P/R/F1; loc/qty: exact-match). For future predictions, reuse Steps 3–4 with \emph{model} set to $\theta^\ast$. \\
\Return $\theta^\ast$.\\
\end{algorithm}

\section{Experiments setup and Preparation} 
\label{sec: Experiments and Analysis}
This section provides a detailed description of the experimental setup, dataset preparation, evaluation metrics, and baseline comparisons used to validate the effectiveness of PVAL for detecting localizing, and quantifying solar panels in satellite imagery.

\subsection{Experiment Setup}

% \sout{
% To rigorously assess the PVAL framework, we conducted experiments in two computational environments:  The first environment utilized OpenAI’s platform for fine-tuning the GPT-4o-2024-08-06 model, allowing direct access to the service’s proprietary architecture and pre-trained capabilities. In parallel, we employed a high-performance computing server equipped with NVIDIA A100-SXM4-80GB GPUs to train and evaluate baseline models and perform additional analyses. 
% }

To rigorously assess the PVAL framework, we conducted experiments in two complementary computational environments. The first environment utilized OpenAI’s platform for fine-tuning the GPT-4o-2024-08-06 model, {\color{black}which provided direct access to the proprietary multimodal architecture and its pre-trained capabilities through the official API.} In parallel, we employed the Sol supercomputer at Arizona State University \cite{jennewein2023sol}, a high-performance computing cluster designed for large-scale AI research. {\color{black}The Sol environment featured NVIDIA A100 GPUs with 80 GB of memory (CUDA 12.7) and dual AMD EPYC 7413 processors with 48 CPU cores running at 3.6 GHz, supported by a Linux operating system (kernel 4.18). Each job was executed on compute nodes with sufficient GPU memory to process large batches of rooftop imagery efficiently. All baseline models and additional simulations were implemented in Python (version 3.9) using the PyTorch framework, with development and debugging conducted in the PyCharm IDE. This setup ensured both high computational throughput and reproducibility across experiments.}

{\color{black}
From a computational efficiency perspective, baseline CNN and ViT models required significant training compute on the Sol cluster (approximately 15–40 GPU-hours depending on architecture), whereas prompted PVAL required no local training and fine-tuned PVAL incurred only a one-time cloud-managed training cost. 
Inference with PVAL takes longer per image (about 1–3 seconds via API) compared to CNN/ViT baselines on A100 GPUs (0.2–0.5 seconds). However, PVAL eliminates repeated retraining when expanding to new regions or schema modifications. This trade-off demonstrates reduced overall resource requirements by shifting the cost from local GPU training cycles to a lightweight API-based workflow.}

An important consideration for PVAL was the output modality of the GPT-4o. Although GPT-4o can effectively process imagery, it produces textual descriptions rather than pixel-level segmentation masks or object-coordinate predictions. To utilize the labels, we formulated our annotations and prompts (e.g., Section \ref{sec:Output Standardization}) in a discrete, text-based manner, ensuring that the model’s outputs could be directly compared to human-generated ground truth labels. The textual output format aligns well with LLM-based workflows, providing interpretable results that can be readily integrated into downstream applications, such as powergrid operation tools and decision support systems.

\begin{figure*}[t]
    \centering
    \includegraphics[width=1\linewidth]{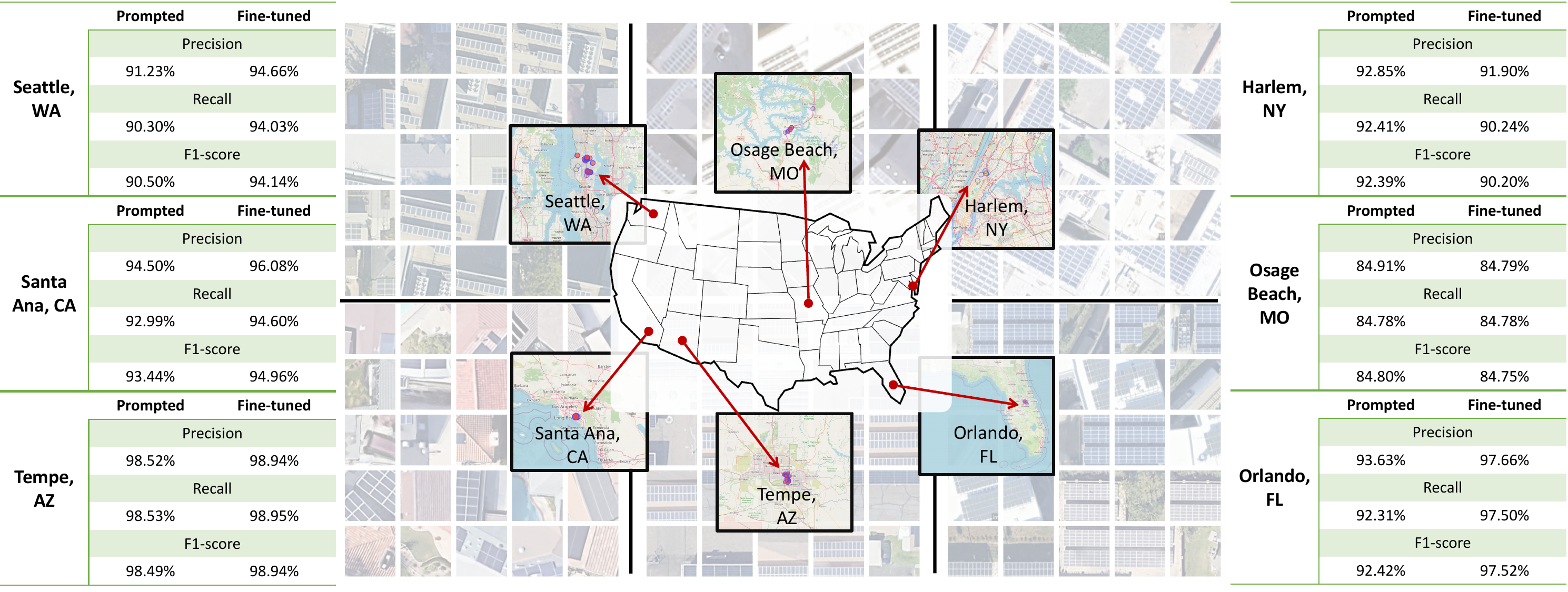}
    \caption{\textcolor{black}{Sample data from six regions across the United States, with tables displaying the weighted average precision, recall, and F1-score for each region.}}
    \label{fig:US_map}
\end{figure*}

\subsection{Data Materials}
\label{sec:data}
{\color{black}
\textbf{Data sources.}  
Our dataset combines U.S. and global satellite imagery to comprehensively evaluate solar PV detection.
For U.S. regions, we collected data from six representative locations: Seattle (WA), Orlando (FL), Osage Beach (MO), Harlem (NY), Tempe (AZ), and Santa Ana (CA).
To assess cross-continental generalization, we further extended the study to six globally distributed cities spanning diverse climate zones and urban morphologies: Sydney (Australia), Cape Town (South Africa), Kuwait City (Middle East), Oxford (Europe), São Paulo (South America), and Shanghai (Asia).
Geographic coordinates of PV installations were obtained from OpenStreetMap (OSM) via the Overpass API, and high-resolution imagery was retrieved from the Google Maps Static API.

\textbf{Transparency of data.} Existing methods often lack transparency regarding their underlying algorithms or training datasets, as their annotated PV datasets are not publicly available.
By contrast, our dataset is constructed entirely from openly accessible sources—geographic coordinates from OSM and imagery from Google Maps—making the collection process transparent and reproducible.
Although annotation requires manual effort, the underlying imagery is straightforward to obtain, ensuring that our dataset can be independently replicated and extended by other researchers.

\textbf{Data structure.}  
Each rooftop image was divided into a $4 \times 4$ grid, producing 16 tiles per rooftop.  
This tiling strategy expanded the dataset size while preserving spatial resolution, ensuring that small-scale PV installations remained visible.  
All tiles were labeled with the binary field \emph{solar panels present}.  
A subset of $10,000$ tiles was further annotated with detailed labels for \emph{location} and \emph{quantity}.  
The location label follows a predefined schema including \emph{top}, \emph{bottom}, \emph{left}, \emph{right}, \emph{center}, and quadrant-based categories (e.g., \emph{top-left}, \emph{bottom-right}).  
The quantity label records the estimated range of panel counts using discrete intervals: $(0,1]$, $(1,5]$, $(5,10]$, and $[10,+\infty)$. 
As illustrated in Figure~\ref{fig:data_structure}, these structured annotations provide a consistent representation of rooftop PV characteristics.

\begin{figure}[t]
    \centering
    \includegraphics[width=0.75\linewidth]{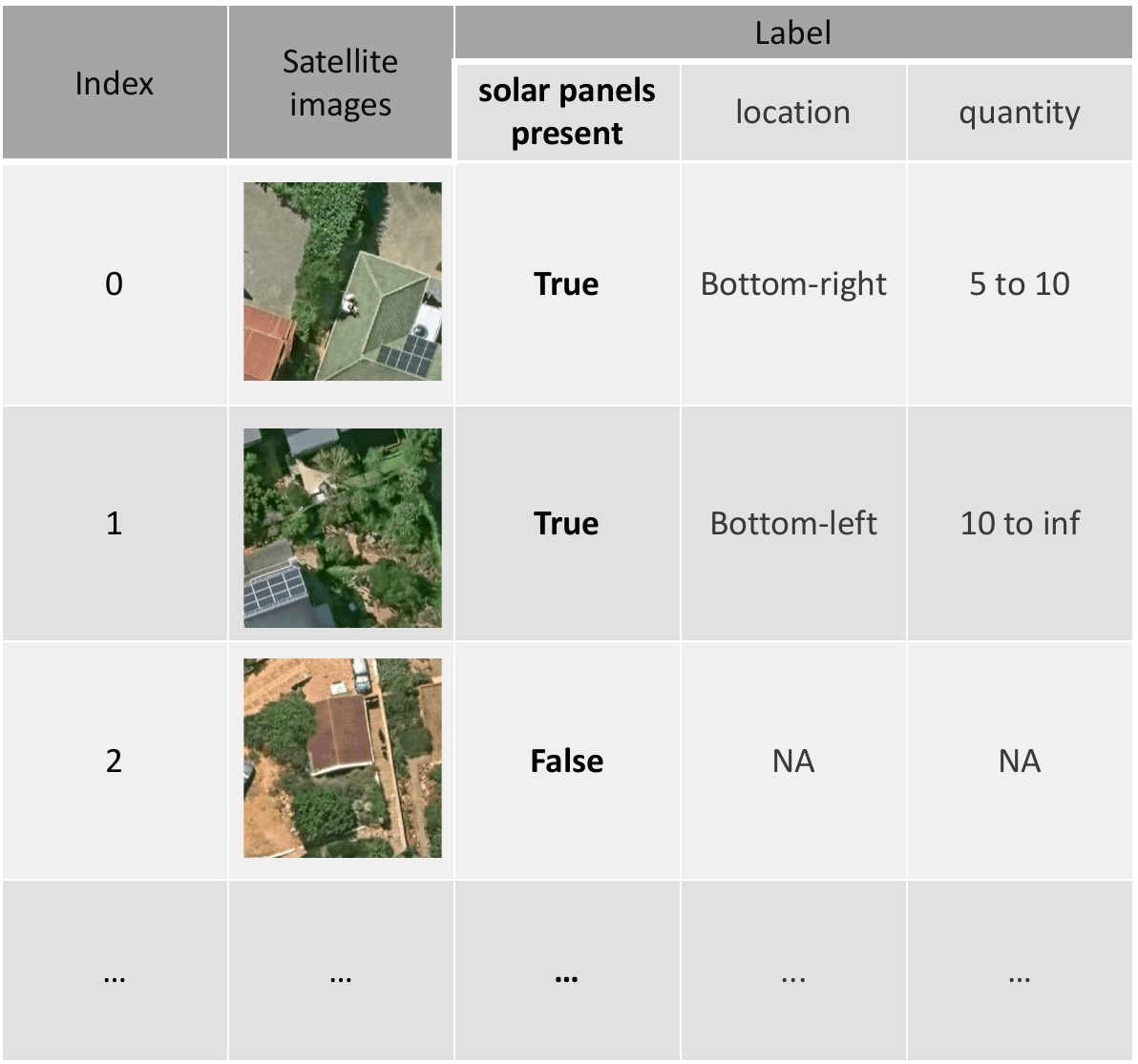}
    \color{black}
    \caption{Illustration of the dataset structure. Each rooftop image is divided into tiles and labeled with solar panel presence, location and quantity information. Example entries are shown with corresponding annotations.}
    \label{fig:data_structure}
\end{figure}

\textbf{Dataset summary.}  
Table~\ref{tab:data-summary} provides an overview of the datasets.  
For consistency, both baseline models (VGG, ResNet, U-Net, ViT, etc.) and our proposed PVAL framework were trained on the same fine-tuning dataset, which consists of 2,000 annotated rooftop tiles from Santa Ana, CA.  
The large-scale evaluation set contains approximately 100,000 tiles from Tempe and Santa Ana.  
Additional datasets from the remaining U.S. and international regions, each with 480 tiles, were used to assess cross-regional and cross-continental generalization.  

\begin{table}[htbp]
  \centering
\textcolor{black}{
  \caption{Summary of training and evaluation datasets. All baseline models use the same fine-tuning data as PVAL.}
  \label{tab:data-summary}
  \resizebox{0.8\linewidth}{!}{%
  \begin{tabular}{lcc}
    \hline
    \textbf{Region / City} & \textbf{Role} & \textbf{Size} \\
    \hline
    Santa Ana, CA & Training (fine-tuning) & 2,000 tiles \\
    Tempe, AZ + Santa Ana, CA & Large-scale Test & $\sim$100,000 tiles \\
    Seattle, WA & Cross-regional Test & 480 tiles \\
    Orlando, FL & Cross-regional Test & 480 tiles \\
    Osage Beach, MO & Cross-regional Test & 480 tiles \\
    Harlem, NY & Cross-regional Test & 480 tiles \\
    Sydney, Australia & Cross-continental Test & 480 tiles \\
    Cape Town, South Africa & Cross-continental Test & 480 tiles \\
    Kuwait City, Middle East & Cross-continental Test & 480 tiles \\
    Oxford, Europe & Cross-continental Test & 480 tiles \\
    São Paulo, South America & Cross-continental Test & 480 tiles \\
    Shanghai, Asia & Cross-continental Test & 480 tiles \\
    \hline
  \end{tabular}}
  }
\end{table}
}

% \subsection{Benchmark Models and Evaluation Metrics}
\subsection{Benchmark Models}
To compare the proposed PVAL against established methods, we evaluated a range of benchmark models spanning traditional machine learning techniques, convolutional neural networks (CNNs), and transformer-based architectures. 
Traditional ML approaches such as SVM, Decision Trees, Random Forest, and Logistic Regression utilized Histogram of Oriented Gradients (HOG) features.
For CNN-based baselines, we fine-tuned U-Net \cite{bouazizintegrated}, ResNet-152 \cite{he2016deep}, VGG-19 \cite{simonyan2014very}, and Inception-v3 \cite{szegedy2016rethinking}, adapting their output layers for binary classification. These models leverage deep feature hierarchies for robust representation learning.
We also evaluated a Vision Transformer (ViT-Base-16 \cite{dosovitskiy2020image}), which is pre-trained on ImageNet-21k and processes images as patches through transformer layers. Together, these baseline methods provide a comprehensive reference point for assessing the effectiveness of our proposed LLM-based approach.

{\color{black}
Unlike these benchmark models, PVAL goes beyond single-task classification or dense segmentation by unifying multiple solar assessment tasks: detection, localization, and quantification, within a single schema-guided framework. Moreover, by fine-tuning a multimodal LLM, PVAL produces structured JSON outputs enriched with likelihood and confidence metrics, enabling seamless integration into GIS/EMS platforms and supporting deployment grade applications such as PV inventory management and hosting-capacity analysis. 
% This design provides transparency, adaptability, and uncertainty-awareness not achievable with conventional CNN, U-Net, or ViT approaches.  
}

{\color{black}
\subsection{Benchmark Model Configuration and Hyperparameter Tuning}
We employed five backbone architectures: Inception-v3, VGG19, ResNet-152, a Vision Transformer (ViT), and a U-Net encoder classifier. In every case, the final classification head was replaced with a fully connected layer producing one logit for binary prediction (solar vs. no solar). All models were trained on the same dataset splits, with identical preprocessing, hardware resources, and training conditions to ensure fair comparison. Training or fine-tuning was performed for 10 epochs using binary cross-entropy loss with the Adam optimizer, a fixed learning rate of $10^{-3}$, and a batch size of 16. For evaluation, we restored the checkpoint that achieved the best validation accuracy.

\textbf{Inception-v3} begins with a seven-by-seven convolution with a stride of two, followed by max pooling, a three-by-three convolution, and another max pooling. The feature extractor is organized into Inception modules: three Inception-A modules (one-by-one, three-by-three, and five-by-five convolutions with pooling), five Inception-B modules (factorized one-by-seven and seven-by-one convolutions), and two Inception-C modules with wider filters. After global average pooling, the original classifier (2048–1000) was replaced by 2048–1, and the auxiliary branch was disabled.

\textbf{ResNet-152} starts with a seven-by-seven convolution and max pooling, followed by four residual stages. Each stage consists of bottleneck blocks with one-by-one, three-by-three, and one-by-one convolutions plus identity skips. The four stages contain 3, 8, 36, and 3 bottleneck blocks. After global average pooling, the original classifier (2048–1000) was replaced by 2048–1.

\textbf{VGG19} is arranged into five convolutional blocks. The first two contain two three-by-three convolutions each, while the last three contain four three-by-three convolutions. Each block is followed by max pooling. The convolutional stack feeds into three fully connected layers (4096, 4096, 1000). We modified the final layer to map 4096–1.

\textbf{ViT-Base-16} was constructed without pretraining using a lightweight configuration. Input images (224×224) were divided into two-by-two patches, producing 112×112 tokens. Each token was projected to a 64-dimensional embedding, with a class token prepended and positional embeddings added. The encoder contained three Transformer layers with four self-attention heads and feed-forward dimension 1028. The class token output was passed through a linear layer (64–1).

\textbf{U-Net} encoder classifier was adapted from the contracting path of U-Net. It began with a double convolution mapping three channels to 64, followed by three downsampling blocks (64–128, 128–256, 256–512). Each block used two-by-two max pooling followed by two three-by-three convolutions with batch normalization and ReLU. The encoder output was reduced with global average pooling and fed to a fully connected layer (512–1).
}

{\color{black}
\subsection{LLM Fine-Tuning and Inference}
The fine-tuning process was implemented through OpenAI’s official API infrastructure \footnote{\url{https://platform.openai.com/docs/api-reference/fine-tuning}}. Labeled training data were converted into JSONL format, where each entry contained a task description, the base64-encoded rooftop image, and the expected output in a standardized JSON schema specifying whether solar panels were present, their location, and their estimated quantity. This format follows OpenAI’s requirements for fine-tuning multimodal models and ensures that both the inputs and outputs have a consistent structure that the model can learn to reproduce.

The JSONL training file was uploaded to OpenAI’s servers through the file upload interface of the API, with its purpose designated as fine-tuning. A fine-tuning job was then initiated by specifying the base model (GPT-4o), the uploaded training file, and the number of training epochs, which we set to five. During fine-tuning, the pretrained GPT-4o weights were further optimized on our solar classification dataset using supervised learning. The optimization objective was cross-entropy loss between the model’s generated tokens and the ground-truth JSON schema in each training example. In effect, the model was encouraged not only to detect the presence of solar panels from the rooftop image, but also to place its response in the exact JSON fields that encode presence, location, and quantity. This structured learning reduces ambiguity at inference time and enforces output regularity, which is important for downstream evaluation and large-scale automation.

The progress of each fine-tuning job was monitored through the API until completion, at which point a unique identifier for the fine-tuned model was returned. After training, this identifier was used for inference with the same API. During inference, rooftop images were again provided in base64 format together with the classification prompt, and the fine-tuned model returned structured JSON outputs consistent with the schema used in training. Because the training schema and inference schema were aligned, the model generalized reliably to unseen images and enabled large-scale automated prediction.
}

\begin{figure*}[h!]
    \centering
    \includegraphics[width=0.9\linewidth]{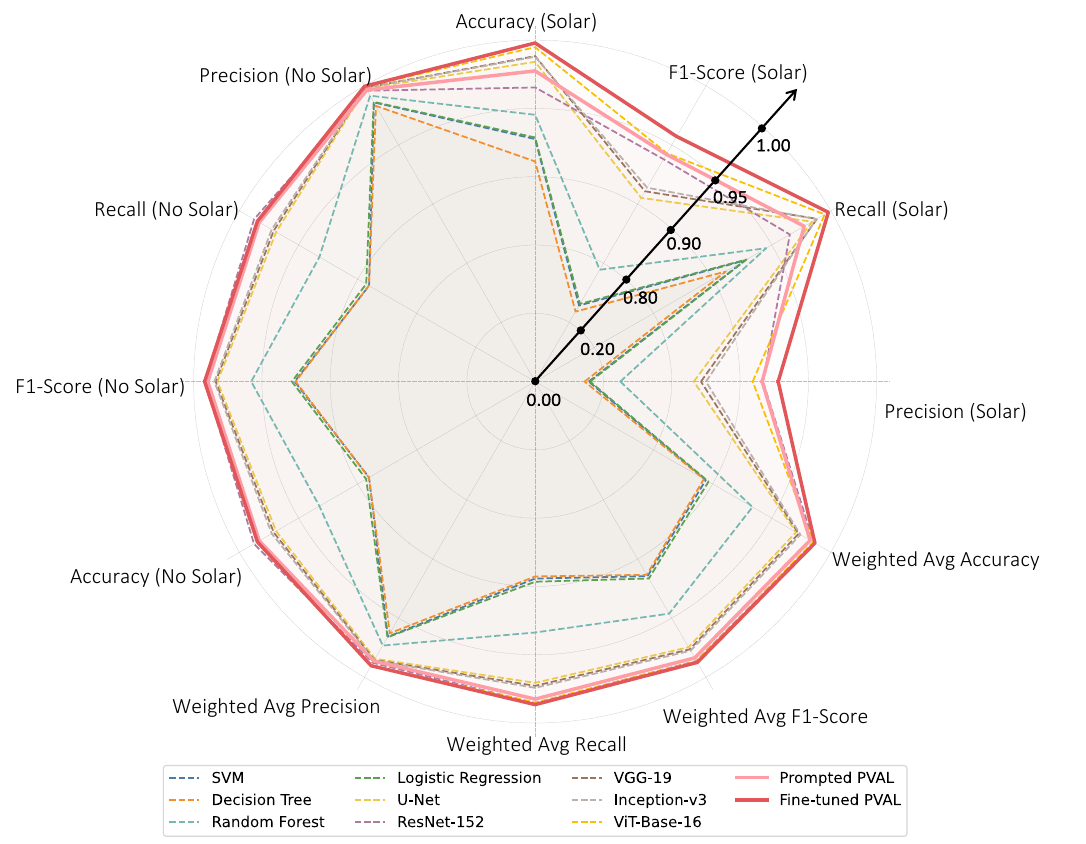}
    \caption{Radar chart comparing the performance of various models on solar panel detection. Metrics include precision, recall, F1-score, and accuracy for "No Solar," "Solar," and weighted averages. "Prompted PVAL" and "Fine-tuned PVAL" are highlighted in pink and red, respectively, to emphasize their performance.}
    \label{fig:Models_performance}
\end{figure*}

\subsection{Evaluation Metrics}
\label{sec:Evaluation Metrics}
\textcolor{black}{Since our framework produces structured natural language and JSON outputs rather than pixel-level masks, conventional vision metrics such as IoU or mAP are not directly applicable. Instead, we adopt precision, recall, F1-score, and accuracy, which align naturally with schema-guided outputs for presence, location, and quantity.}
% To comprehensively assess the performance of the proposed method and other benchmarks, we employed the following evaluation metrics:

\noindent \textbf{Precision:}  
The ratio of true positive detections (\(TP\)) to the total number of positive detections (\(TP + FP\)), measuring the accuracy of the model’s positive predictions. It is defined as:  
\begin{equation}
    \text{Precision} = \frac{TP}{TP + FP}
\end{equation}

\noindent \textbf{Recall:}  
The ratio of true positive detections (\(TP\)) to the total number of actual positives (\(TP + FN\)), reflecting the model’s ability to identify all instances of solar panels. It is defined as:  
\begin{equation}
\text{Recall} = \frac{TP}{TP + FN}
\end{equation}
\noindent \textbf{F1 Score:} 
The harmonic mean of precision and recall, providing a balanced measure of the model’s overall performance. F1-Score is defined as:  
\begin{equation}
\text{F1 Score} = 2 \cdot \frac{\text{Precision} \cdot \text{Recall}}{\text{Precision} + \text{Recall}}
\end{equation}
\noindent \textbf{Accuracy of Location and Quantity:}  
Since the model generates location or quantity descriptions (e.g., \textit{``left"}, \textit{``top-left"}, or \textit{``10 to inf"}), the accuracy metric is defined as the percentage of descriptions that are \textbf{exactly correct}, calculated using the formula:
\begin{equation}
\text{Accuracy} = \frac{\text{Number of Correct Descriptions}}{\text{Total Number of Descriptions}}
\end{equation}

\section{Simulation Results and Analysis}
\subsection{Result Analysis for Prompting Strategies}
\label{sec: Prompting}

\begin{table*}[t]
\centering
\caption{PVAL Prompt vs PVAL Fine-tuned for Solar Panel Detection Across Regions}
\renewcommand{\arraystretch}{1.3}
\setlength{\tabcolsep}{4pt}
\resizebox{1\textwidth}{!}{
\begin{tabular}{l|l|cccc|cccc|cccc}
\hline
\multirow{2}{*}{\textbf{Methods}} & \multirow{2}{*}{\textbf{Region}} & \multicolumn{4}{c|}{\textbf{Solar}} & \multicolumn{4}{c|}{\textbf{No Solar}} & \multicolumn{4}{c}{\textbf{Weighted Average}} \\
\cline{3-14}
 & & \textbf{Precision} & \textbf{Recall} & \textbf{F1-Score} & \textbf{Accuracy} & \textbf{Precision} & \textbf{Recall} & \textbf{F1-Score} & \textbf{Accuracy} & \textbf{Precision} & \textbf{Recall} & \textbf{F1-Score} & \textbf{Accuracy} \\
\hline
\multirow{6}{*}{PVAL Prompt} & Santa Ana, CA & 66.48\% & 90.98\% & 76.82\% & 90.98\% & 98.60\% & 93.29\% & 95.87\% & 93.29\% & 94.50\% & 92.99\% & 93.44\% & 92.99\% \\
 & Seattle, WA & 78.72\% & 92.50\% & 85.06\% & 92.50\% & 96.55\% & 89.36\% & 92.82\% & 89.36\% & 91.23\% & 90.30\% & 90.50\% & 90.30\% \\
 & Tempe, AZ & 98.63\% & \textbf{\colorbox{green!10}{99.77\%}} & 99.20\% & \textbf{\colorbox{green!10}{99.77\%}} & 96.55\% & 86.05\% & 91.36\% & 86.05\% & 98.52\% & 98.53\% & 98.49\% & 98.53\% \\
 & Orlando, FL & 82.86\% & \textbf{\colorbox{green!10}{100.00\%}} & 90.62\% & \textbf{\colorbox{green!10}{100.00\%}} & \textbf{\colorbox{green!10}{100.00\%}} & 87.76\% & 93.48\% & 87.76\% & 93.63\% & 92.31\% & 92.42\% & 92.31\% \\
 & Osage Beach, MO & 81.82\% & \textbf{\colorbox{green!10}{85.71\%}} & \textbf{\colorbox{green!10}{83.72\%}} & \textbf{\colorbox{green!10}{85.71\%}} & \textbf{\colorbox{green!10}{87.50\%}} & 84.00\% & 85.71\% & 84.00\% & 
 \textbf{\colorbox{green!10}{84.91\%}} &
 \textbf{\colorbox{green!10}{84.78\%}} & \textbf{\colorbox{green!10}{84.80\%}} & \textbf{\colorbox{green!10}{84.78\%}} \\
 & Harlem, NY & \textbf{\colorbox{green!10}{88.37\%}} & 97.44\% & \textbf{\colorbox{green!10}{92.68\%}} & 97.44\% & 97.22\% & \textbf{\colorbox{green!10}{87.50\%}} & \textbf{\colorbox{green!10}{92.11\%}} & \textbf{\colorbox{green!10}{87.50\%}} & \textbf{\colorbox{green!10}{92.85\%}} & \textbf{\colorbox{green!10}{92.41\%}} & \textbf{\colorbox{green!10}{92.39\%}} & \textbf{\colorbox{green!10}{92.41\%}} \\
\hline
\multirow{6}{*}{PVAL Fine-tuned} & Santa Ana, CA & \textbf{\colorbox{orange!10}{71.18\%}} & \textbf{\colorbox{orange!10}{99.15\%}} & \textbf{\colorbox{orange!10}{82.87\%}} & \textbf{\colorbox{orange!10}{99.15\%}} & \textbf{\colorbox{orange!10}{99.86\%}} & \textbf{\colorbox{orange!10}{93.91\%}} & \textbf{\colorbox{orange!10}{96.79\%}} & \textbf{\colorbox{orange!10}{93.91\%}} & \textbf{\colorbox{orange!10}{96.08\%}} & \textbf{\colorbox{orange!10}{94.60\%}} & \textbf{\colorbox{orange!10}{94.96\%}} & \textbf{\colorbox{orange!10}{94.60\%}} \\
 & Seattle, WA & \textbf{\colorbox{orange!10}{84.78\%}} & \textbf{\colorbox{orange!10}{97.50\%}} & \textbf{\colorbox{orange!10}{90.70\%}} & \textbf{\colorbox{orange!10}{97.50\%}} & \textbf{\colorbox{orange!10}{98.86\%}} & \textbf{\colorbox{orange!10}{92.55\%}} & \textbf{\colorbox{orange!10}{95.60\%}} & \textbf{\colorbox{orange!10}{92.55\%}} & \textbf{\colorbox{orange!10}{94.66\%}} & \textbf{\colorbox{orange!10}{94.03\%}} & \textbf{\colorbox{orange!10}{94.14\%}} & \textbf{\colorbox{orange!10}{94.03\%}} \\
 & Tempe, AZ & \textbf{\colorbox{orange!10}{99.08\%}} & \textbf{\colorbox{orange!10}{99.77\%}} & \textbf{\colorbox{orange!10}{99.42\%}} & \textbf{\colorbox{orange!10}{99.77\%}} & \textbf{\colorbox{orange!10}{97.62\%}} & \textbf{\colorbox{orange!10}{91.11\%}} & \textbf{\colorbox{orange!10}{94.25\%}} & \textbf{\colorbox{orange!10}{91.11\%}} & \textbf{\colorbox{orange!10}{98.94\%}} & \textbf{\colorbox{orange!10}{98.95\%}} & \textbf{\colorbox{orange!10}{98.94\%}} & \textbf{\colorbox{orange!10}{98.95\%}} \\
 & Orlando, FL & \textbf{\colorbox{orange!10}{93.55\%}} & \textbf{\colorbox{orange!10}{100.00\%}} & \textbf{\colorbox{orange!10}{96.67\%}} & \textbf{\colorbox{orange!10}{100.00\%}} & \textbf{\colorbox{orange!10}{100.00\%}} & \textbf{\colorbox{orange!10}{96.08\%}} & \textbf{\colorbox{orange!10}{98.00\%}} & \textbf{\colorbox{orange!10}{96.08\%}} & \textbf{\colorbox{orange!10}{97.66\%}} & \textbf{\colorbox{orange!10}{97.50\%}} & \textbf{\colorbox{orange!10}{97.52\%}} & \textbf{\colorbox{orange!10}{97.50\%}} \\
 & Osage Beach, MO & \textbf{\colorbox{orange!10}{85.00\%}} & 80.95\% & 82.93\% & 80.95\% & 84.62\% & \textbf{\colorbox{orange!10}{88.00\%}} & \textbf{\colorbox{orange!10}{86.27\%}} & \textbf{\colorbox{orange!10}{88.00\%}} & 84.79\% & 
 \textbf{\colorbox{orange!10}{84.78\%}}
  & 84.75\% & \textbf{\colorbox{orange!10}{84.78\%}} \\
 & Harlem, NY & 82.98\% & \textbf{\colorbox{orange!10}{100.00\%}} & 90.70\% & \textbf{\colorbox{orange!10}{100.00\%}} & \textbf{\colorbox{orange!10}{100.00\%}} & 81.40\% & 89.74\% & 81.40\% & 91.90\% & 90.24\% & 90.20\% & 90.24\%\\
\hline
\end{tabular}
}
\label{tab:results_gpt}
\end{table*}

The first part of Table \ref{tab:results_gpt} presents the performance of PVAL with prompt strategies for solar panel detection across various regions.
% against a fine-tuned GPT-4o model
The class ``Solar" indicates images containing solar panels, while the class ``No Solar" indicates images without solar panels.

Notably, existing open-source models, such as Vision Transformer (ViT), Inception-v3, and other large pre-trained architectures, face significant challenges when applied to domain-specific tasks like solar panel detection. For example, ViT-Base-16, pre-trained on ImageNet-21k, is optimized for general object recognition and lacks exposure to domain-specific objects such as solar panels. As a result, these models fail to produce meaningful results in solar panel detection tasks without extensive fine-tuning and domain adaptation, which require considerable computational resources and specialized datasets. This limitation renders them impractical for organizations or researchers with limited access to these resources.

Unlike the traditional pre-trained vision model, the PVAL can adapt to solar panel detection directly by leveraging contextual understanding and task-specific prompts. 
We use a green background to mark the values where prompted PVAL can achieve the same or better performance compared to the fine-tuned PVAL.
Notably, prompted PVAL achieves competitive average accuracy and F1-scores in regions such as Osage Beach, MO, and Harlem, NY, without requiring fine-tuning or additional training. A concise comparison can be seen in Figure \ref{fig:US_map}. These findings underscore its potential as a cost-effective and efficient solution for organizations seeking robust performance in solar panel detection without incurring the high costs associated with data collection, annotation, and model fine-tuning.

{\color{black}
While fine-tuned PVAL achieves superior overall performance, a slight reduction in precision is observed for the solar category in Harlem, NY, compared to prompted PVAL (Table~\ref{tab:results_gpt}). This can be explained by the urban context of Harlem, where dense rooftop layouts and shadowed conditions increase the likelihood of false positives. The fine-tuned model prioritizes higher recall under such conditions, ensuring that true solar installations are less likely to be overlooked, albeit at the expense of marginally reduced precision.
}

\subsection{Result Analysis for Fine-tuned Models}
\label{sec: Fine-tuned}

\begin{table*}[h!]
\centering
\caption{Comprehensive Results Table for Solar Panel Detection Methods after fine-tuning or training}
\renewcommand{\arraystretch}{1.3} % Adjust row height for better readability
\setlength{\tabcolsep}{4pt} % Adjust column spacing
\resizebox{1\textwidth}{!}{
\begin{tabular}{c|l|cccc|cccc|cccc}
\hline
\multirow{2}{*}{\textbf{}} & \multirow{2}{*}{\centering \textbf{Methods}} & \multicolumn{4}{c|}{\textbf{Solar}} & \multicolumn{4}{c|}{\textbf{No Solar}} & \multicolumn{4}{c}{\textbf{Weighted Average}} \\
\cline{3-14}
 & & \textbf{Precision} & \textbf{Recall} & \textbf{F1-Score} & \textbf{Accuracy} & \textbf{Precision} & \textbf{Recall} & \textbf{F1-Score} & \textbf{Accuracy} & \textbf{Precision} & \textbf{Recall} & \textbf{F1-Score} & \textbf{Accuracy} \\
\hline
\multirow{4}{*}{Small Size Models} & SVM & 15.70\% & 71.01\% & 25.72\% & 71.01\% & 94.40\% & 56.16\% & 70.42\% & 56.16\% & 86.28\% & 57.69\% & 65.81\% & 57.69\% \\
 & Decision Tree & 14.48\% & 64.46\% & 23.65\% & 64.46\% & 93.22\% & 56.22\% & 70.14\% & 56.22\% & 85.10\% & 57.07\% & 65.35\% & 57.07\% \\
 & Random Forest & 24.94\% & 78.12\% & 37.81\% & 78.12\% & 96.67\% & 72.96\% & 83.15\% & 72.96\% & 89.27\% & 73.49\% & 78.48\% & 73.49\% \\
 & Logistic Regression & 16.09\% & 71.48\% & 26.27\% & 71.48\% & 94.57\% & 57.14\% & 71.24\% & 57.14\% & 86.48\% & 58.62\% & 66.60\% & 58.62\% \\
\hline
\multirow{6}{*}{Large Size Models} & U-Net \cite{bouazizintegrated} & 46.47\% & 93.64\% & 62.11\% & 93.64\% & 99.17\% & 87.59\% & 93.02\% & 87.59\% & 93.74\% & 88.22\% & 89.83\% & 88.22\% \\
 & ResNet-152 \cite{he2016deep} & 66.74\% & 86.13\% & 75.20\% & 86.13\% & 98.35\% & \textbf{\colorbox{blue!10}{95.06\%}} & 96.68\% & \textbf{\colorbox{blue!10}{95.06\%}} & 95.09\% & 94.14\% & 94.46\% & 94.14\% \\
 & VGG-19 \cite{simonyan2014very} & 48.56\% & 95.32\% & 64.35\% & 95.32\% & 99.40\% & 88.39\% & 93.57\% & 88.39\% & 94.15\% & 89.10\% & 90.55\% & 89.10\% \\
 & Inception-v3 \cite{szegedy2016rethinking} & 50.03\% & 95.11\% & 65.57\% & 95.11\% & 99.37\% & 89.08\% & 93.94\% & 89.08\% & 94.28\% & 89.70\% & 91.02\% & 89.70\% \\
 & ViT-Base-16 \cite{dosovitskiy2020image} & 63.71\% & 97.91\% & 77.19\% & 97.91\% & 99.74\% & 93.59\% & 96.57\% & 93.59\% & 96.03\% & 94.03\% & 94.57\% & 94.03\% \\
 \cline{2-14}
 & PVAL Fine-tuned & 
 \colorbox{red!10}{\textbf{71.18\%}} & 
 \colorbox{red!10}{\textbf{99.15\%}} & 
 \colorbox{red!10}{\textbf{82.87\%}} & 
 \colorbox{red!10}{\textbf{99.15\%}} & \colorbox{red!10}{\textbf{99.86\%}} & 
 93.91\% & 
 \colorbox{red!10}{\textbf{96.79\%}} & 
 93.91\% & 
 \textbf{\colorbox{red!10}{96.08\%}} & \textbf{\colorbox{red!10}{94.60\%}} & \textbf{\colorbox{red!10}{94.96\%}} & \textbf{\colorbox{red!10}{94.60\%}} \\
\hline
\end{tabular}
}
\label{tab:res_all}
\end{table*}

The fine-tuned PVAL demonstrates superior performance compared to the prompted PVAL across most metrics and geographical regions, as shown in Table \ref{tab:results_gpt}. The fine-tuned model achieves the highest accuracy in five out of six regions studied.
Fine-tuning significantly improves precision, recall, and F1-scores for solar panel detection across multiple regions, including Santa Ana (CA), Seattle (WA), Tempe (AZ), and Orlando (FL). For instance, the fine-tuned model achieved a F1-score of 97.52\% in Orlando, compared to 92.42\% by the prompted approach, representing a clear advantage.
% \begin{table}[htbp]
%   \centering
%   \color{black}
%   \caption{Test Cross-Entropy Loss with Variance}
%   \label{tab:ce-loss-var}
%   \resizebox{\linewidth}{!}{%
%   \begin{tabular}{l|cccccccccc}
%     \hline
%     Method & SVM & DecTree & RandForest & LogReg & U-Net & ResNet152 & VGG19 & Inc-v3 & ViT-B16 & PVAL \\
%     \hline
%     CE Loss (mean ± std) & 0.68 ± 0.07 & 0.69 ± 0.08 & 0.56 ± 0.05 & 0.67 ± 0.07 & 0.36 ± 0.04 & 0.23 ± 0.11 & 0.35 ± 0.03 & 0.34 ± 0.09 & 0.24 ± 0.12 & \textbf{0.12 ± 0.02} \\
%     \hline
%   \end{tabular}}
% \end{table}
\begin{table}[htbp]
  \centering
  \color{black}
  \caption{Test Cross-Entropy Loss with Variance}
  \label{tab:ce-loss-var}
  \resizebox{\linewidth}{!}{%
  \begin{tabular}{l|ccccc}
    \hline
    Method & SVM & DecTree & RandForest & LogReg & U\hbox{-}Net \\
    \hline
    CE Loss (mean $\pm$ std) & 0.68 $\pm$ 0.07 & 0.69 $\pm$ 0.08 & 0.56 $\pm$ 0.05 & 0.67 $\pm$ 0.07 & 0.36 $\pm$ 0.04 \\
    \hline\hline
    Method & ResNet152 & VGG19 & Inc\hbox{-}v3 & ViT\hbox{-}B16 & \textbf{PVAL} \\
    \hline
    CE Loss (mean $\pm$ std) & 0.23 $\pm$ 0.11 & 0.35 $\pm$ 0.03 & 0.34 $\pm$ 0.09 & 0.24 $\pm$ 0.12 & \textbf{0.12 $\pm$ 0.02} \\
    \hline
  \end{tabular}}
\end{table}
The fine-tuned PVAL not only surpasses the prompted counterpart but also outperforms traditional benchmark models. We take Santa Ana as an example, as shown in Table \ref{tab:res_all}, highlighting the impact of domain adaptation on the specialized task of solar panel detection.
{
{\color{black} Furthermore, Table~\ref{tab:ce-loss-var} reports the test cross-entropy loss averaged over three independent runs. The fine-tuned PVAL achieves the lowest loss (0.12 ± 0.02), clearly outperforming all baseline methods. This result underscores both the robustness and efficiency of the proposed framework.}
}

The fine-tuned PVAL model achieves remarkable results, including a solar F1-score of 82.87\% and a weighted average F1-score of 94.96\%. Additionally, it attains an impressive solar recall of 99.15\%, demonstrating its ability to detect solar panels with near-perfect completeness. These metrics make fine-tuned PVAL the best-performing model, showcasing the effectiveness of leveraging a fine-tuned LLM for solar panel detection tasks.

However, benchmark models such as ResNet-152 and ViT-Base-16 also provide valuable insights into the challenges and trade-offs inherent in solar panel detection. ResNet-152 achieves the highest recall (95.06\%) and accuracy (95.06\%) for the ``No Solar" class among all models. This indicates that ResNet-152 is highly effective at minimizing false negatives for predictions of image without solar panels. The reason for this lies in its deep architecture and hierarchical feature extraction, which enable it to capture subtle differences between solar and non-solar regions. However, its relatively lower performance in the solar category (F1-score of 75.20\%) suggests that the model struggles with balancing class-specific trade-offs, likely due to the lack of task-specific knowledge.

ViT-Base-16 also exhibits strong performance, achieving a F1-score of 77.19\% for the ``solar" class and a weighted average F1-score of 94.57\%. Its ability to handle both ``Solar" and ``No Solar" classes effectively highlights the strength of transformer-based architectures in modeling long-range dependencies and capturing complex spatial features. Nonetheless, ViT-Base-16's performance, while competitive, is slightly inferior to Fine-tuned PVAL, primarily due to the lack of domain-specific adaptation. The reliance on general-purpose features hinders its ability to achieve optimal results in a specialized task like solar panel detection.
% An intuitive comparison can be seen in Figure \ref{fig:Models_performance}. A fine-tuned PVAL combines the benefits of PV-task-specific training with the flexibility of generative pre-trained transformers. Fine-tuning the LLM on solar panel data can achieve both high recall and precision across all categories, striking an effective balance that traditional vision models struggle to achieve. 

Figure \ref{fig:Models_performance} highlights the performance of a fine-tuned PVAL, which combines PV-specific training with the flexibility of generative transformers. Fine-tuning on solar panel data enables high recall and precision across all categories, outperforming all baseline models in achieving this balance.

\subsection{Scalability and adaptability of PVAL}

\begin{figure}[htbp]
    \centering
    \includegraphics[width=1\linewidth]{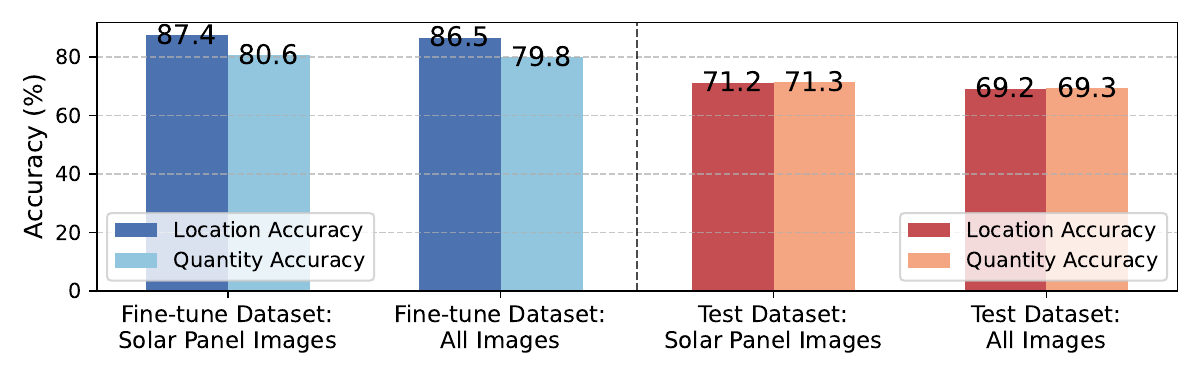}
    \caption{Accuracy of solar panel location and quantity for fine-tune and test datasets.}
    \label{fig:acc_location_quantity}
\end{figure}

In addition to its scalability to multiple regions, as shown in Sections \ref{sec: Prompting} and \ref{sec: Fine-tuned}, we demonstrate PVAL also exhibits adaptability to diverse tasks, including localization and quantification for solar panels.

Figure \ref{fig:acc_location_quantity} highlights the prediction accuracy of the fine-tuned PVAL in identifying both the position and quantity of solar panels. The fine-tuned dataset achieves notably high accuracy, with position prediction at 87.38\% for solar panel images and 86.50\% across all images, reflecting the effectiveness of the model after optimization. Meanwhile, the test dataset maintains reasonable performance, with 71.24\% accuracy for position prediction on solar panel images, underscoring the model's generalizability to unseen data. 
% These results collectively indicate the robustness of the model in handling both localization and numerical estimation tasks. 

\begin{figure*}[htbp]
    \centering
    \includegraphics[width=1\linewidth]{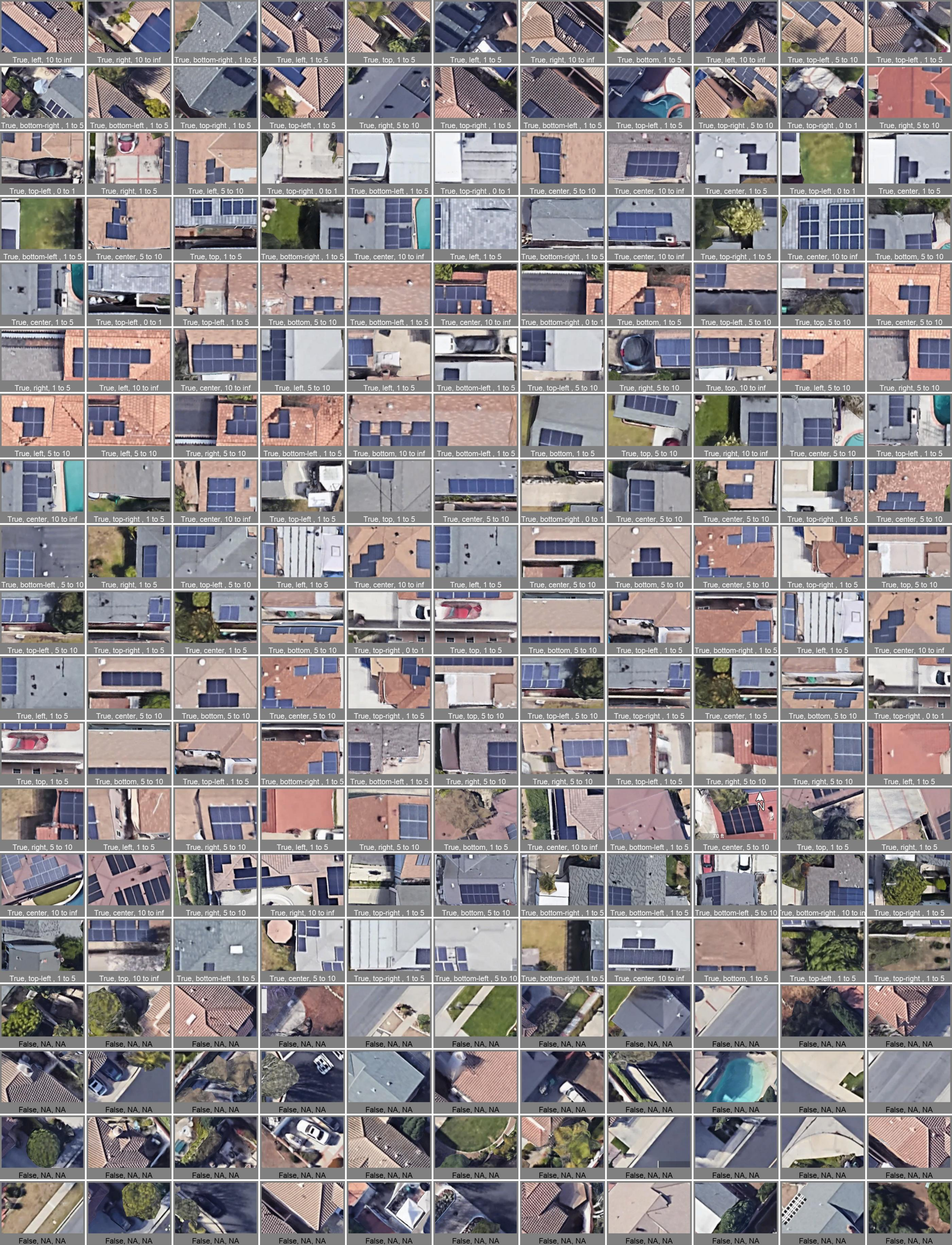}
    \caption{Examples of inferences by fine-tuned PVAL for detecting, localizing, and quantifying solar panels. The results demonstrate the model's precision in identifying and describing the spatial positioning of solar panels in satellite imagery.}
    \label{fig:GPT4o_predict_correct}
\end{figure*}

It is worth noting that the evaluation metrics for location and quantity rely on an ``exact match" criterion (see Section \ref{sec:Evaluation Metrics}).
However, in practical applications, strict accuracy in the \textit{``location"} and \textit{``quantity"} metrics may not always be necessary. For instance, identifying a solar panel's location as \textit{``left"} can be acceptable for a ground truth labeled as \textit{``top-left"}.

To further illustrate this, Figure \ref{fig:GPT4o_predict_correct} provides visual examples of predicted solar panel locations and quantities. The results demonstrate that PVAL effectively identified and described the positions of solar panels using descriptors such as \textit{``bottom-right"}, \textit{``top-left"}, and \textit{``center"}, etc. These predictions closely align with manually annotated ground-truth regions. Moreover, the examples highlight the model's robust spatial reasoning capabilities, even in complex scenarios where solar panels are distributed across varied spatial orientations, including edges, corners, or central areas. This capacity for precise spatial interpretation underscores the model's potential for large-scale solar panel detection tasks.

\subsection{Confidence-Driven Auto-Labeling Mechanism}
\begin{figure}[htbp]
    \centering
    \includegraphics[width=1\linewidth]{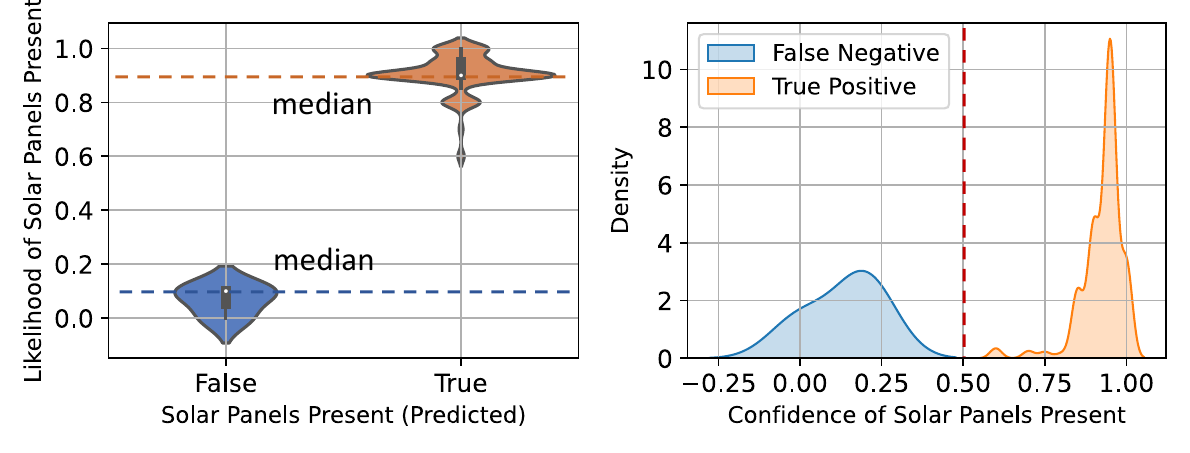}
    \caption{(Left) Violin plot showing the likelihood of solar panels present as predicted for True (``Solar") and False (``No Solar") classifications, along with their respective median values. (Right) Density plot of prediction confidence for True Positive and False Negative cases, generated using Kernel Density Estimation (KDE), with a vertical red line marking the threshold at 0.5.}
    \label{fig:confidence}
\end{figure}

The dual-metric approach of likelihood and confidence offers a comprehensive framework for understanding the model's decision-making process and assessing its reliability in classifying solar panel presence. In this experiment, \textit{likelihood} quantifies the probability of solar panel presence in an image, while \textit{confidence} reflects the model’s internal certainty regarding its classification.

Figure~\ref{fig:confidence} illustrates these metrics from two perspectives. The violin plot on the left visualizes the likelihood distributions for images classified as True (``Solar") and False (``No Solar"). As expected, images correctly classified as True exhibit significantly higher median likelihood values than those labeled False, reaffirming the alignment between likelihood scores and the true presence of solar panels.

The density plot on the right focuses on confidence values for True Positive and False Negative outcomes. True Positive cases cluster tightly around confidence values near 1.0, demonstrating the model's strong assurance when making accurate predictions. Conversely, False Negatives are more prevalent at lower confidence levels, often below a 0.5 threshold indicated by the vertical red line. This trend highlights challenging scenarios where the model underestimates the likelihood of solar panels, leading to missed detections.

The density plot on the right focuses on confidence values for True Positive and False Negative outcomes. True Positive cases cluster tightly around confidence values near 1.0, demonstrating the model's strong assurance when making accurate predictions. Conversely, False Negatives are more prevalent at lower confidence levels, often below a 0.5 threshold indicated by the vertical red line. This trend highlights challenging scenarios where the model underestimates the likelihood of solar panels, leading to missed detections.

By analyzing likelihood and confidence together, this experiment underscores their complementary roles in improving the auto-labeling mechanism. High-likelihood, high-confidence predictions correspond to reliable classifications, suitable for automated labeling with minimal manual intervention. 
To enhance system performance, decision thresholds can be adjusted to balance sensitivity and precision, thereby reducing False Negatives without compromising the model's overall reliability. This adaptive mechanism not only ensures accurate auto-labeling but also facilitates targeted interventions, streamlining the management of extensive solar panel datasets for downstream tasks.

\textcolor{black}{In practice, we estimate that adopting a conservative threshold (likelihood $\geq$ 0.90 and confidence $\geq$ 0.80) would allow approximately 65--70\% of images to be auto-labeled without human review, with over 95\% precision in accepted cases. This reduces the manual annotation workload by more than half. For example, annotating 10,000 images typically requires about 50 annotator-hours. With our auto-labeling policy, the same task requires only about 17 hours, saving 33 hours while maintaining equivalent label quality. Such savings make large-scale dataset curation substantially more feasible and cost-effective.}

\subsection{Ablation Study}
\textcolor{black}{To further clarify the role of each design choice, we conducted a compact ablation study. Starting from the full PVAL framework (task decomposition, output standardization, few-shot prompting, and fine-tuning), we removed one component at a time and measured performance on a subset from Tempe, Arizona. Table~\ref{tab:ablation} summarizes the results. Task decomposition (TD) improves spatial reasoning, particularly for location and quantity accuracy. Without output standardization (OS), the model often generates free-form text, reducing the parse rate and lowering downstream accuracy. Few-shot prompting (FS) stabilizes classification and reduces errors in challenging cases. Fine-tuning (FT) provides the largest overall improvement, and detailed results across six U.S. regions are already reported in Table \ref{tab:results_gpt}. 
Together, these results confirm that each component contributes complementary benefits, and their combination is necessary for robust and scalable solar panel detection.
}
\begin{table}[htbp]
\centering
\caption{Ablation study of PVAL components}
\label{tab:ablation}
\color{black}
\renewcommand{\arraystretch}{1.2}
\begin{adjustbox}{width=\linewidth}
\begin{tabular}{lcccc}
\hline
Setting & Presence Acc. (\%) & Location Acc. (\%) & Quantity Acc. (\%) & Parse Rate (\%) \\
\hline
% Full (TD + OS + FS + FT) & 94.5 & 87.4 & 80.6 & 100 \\
Full (TD + OS + FS) & 94.5 & 87.4 & 80.6 & 100.0 \\
--TD                     & 91.2 & 78.2 & 72.0 & 100.0 \\
--OS                     & 92.4 & 77.5 & 71.3 & 72.4 \\
--FS (k=0)               & 89.6 & 80.1 & 73.5 & 97.5 \\
% --FT (Prompt only)       & 86.7 & 75.4 & 68.9 & 97 \\
\hline
\end{tabular}
\end{adjustbox}
\end{table}

\subsection{Cross-continental Validation for Scalability}
{\color{black}
We extend the study to six globally distributed cities spanning five climate zones and diverse urban morphologies: Sydney (Australia), Cape Town (South Africa), Kuwait City (Middle East), Oxford (Europe), São Paulo (South America), and Shanghai (Asia). This setting stresses the model along multiple, realistic axes of distribution shift: (i) roof form and material (pitched asphalt shingles vs.\ flat concrete vs.\ metal sheet), (ii) background textures and chromatic context (vegetation, sand, asphalt, water tanks, HVAC equipment), (iii) imaging style and resolution differences across providers, and (iv) class prevalence and object scale (dense PV clusters on industrial roofs vs.\ sparse residential adoption). We apply the same inference recipe and thresholds used in U.S. studies—without per-city re-training—to isolate generalization.

\begin{figure}[htbp]
    \centering
    \includegraphics[width=1\linewidth]{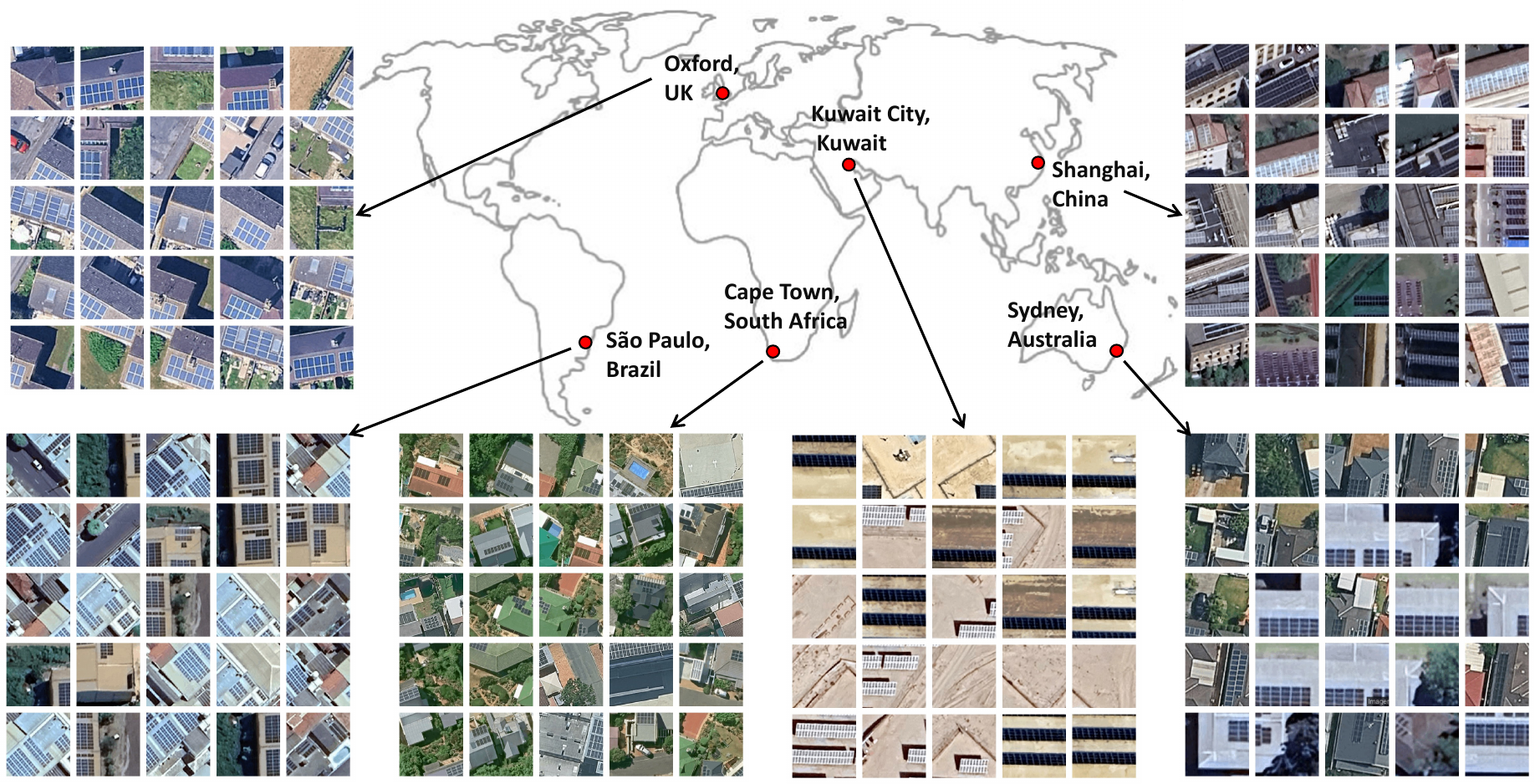}
    \color{black}
    \caption{\textbf{Illustrative global samples and geographic coverage.} 
    Red markers indicate six cities across multiple continents. Around the map we show a visualization of 25 sampled rooftop chips that highlight the variety of roof types, shadows, desert/vegetation backgrounds, and high-rise occlusions.
    }
    \label{fig:Global_Map}
\end{figure}

\begin{figure}[htbp]
    \centering
    \includegraphics[width=1\linewidth]{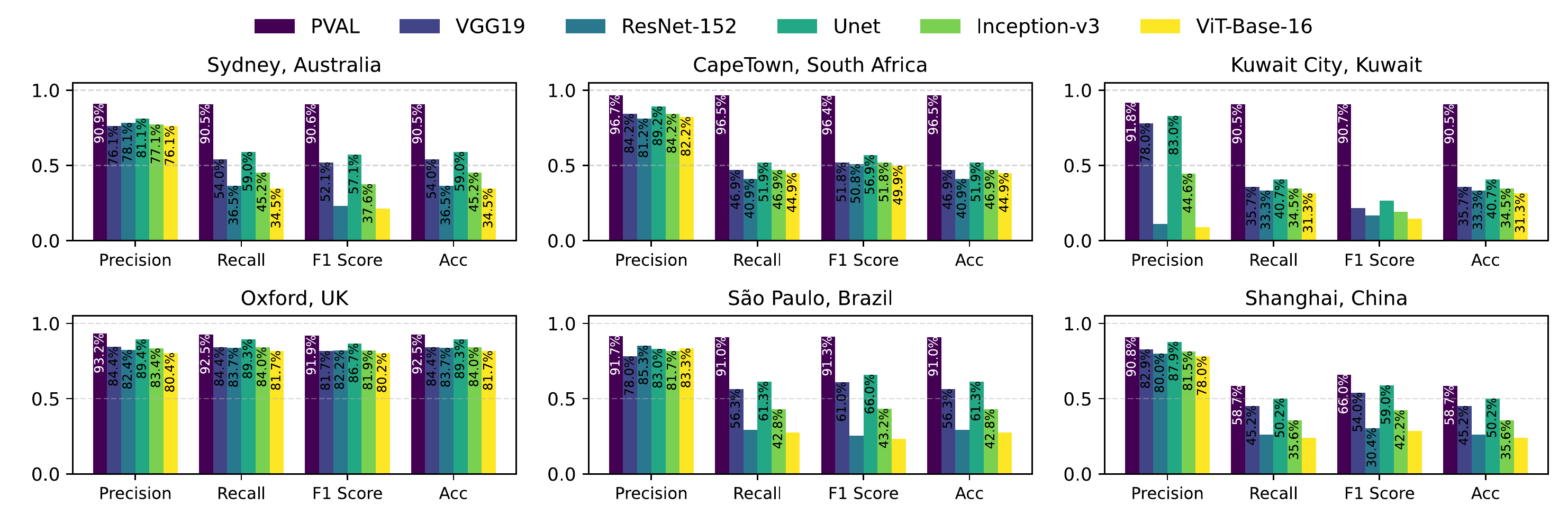}
    \color{black}
    \caption{\textbf{Cross-regional performance across models.} 
    Precision, recall, F1, and accuracy for seven models (PVAL, VGG19, ResNet-152, U-Net, Inception-v3, ViT-Base-16) across the six cities.
    The proposed \textbf{PVAL} consistently ranks top and remains stable across heterogeneous environments, while conventional backbones exhibit larger variance with shifts in roof material, background texture, and illumination. Bars are computed using the same inference settings for all regions to cleanly assess out-of-distribution generalization.}
    \label{fig:Global_results}
\end{figure}

From a global viewpoint, three patterns emerge.
\textbf{(1) Stability under texture/illumination shifts.} Desert glare and high-albedo concrete in Kuwait, and deep shadows in high-rise districts of Shanghai, are challenging for purely appearance-based CNN/ViT baselines; in contrast, PVAL’s schema-guided reasoning and uncertainty-aware scoring sustain balanced precision–recall, limiting false positives from repetitive textures (e.g., skylights, HVAC fins) and false negatives under occlusion.
\textbf{(2) Morphology and scale robustness.} In Cape Town and São Paulo, heterogeneous roof tiling and vegetation background shift the distribution of edge/line cues; PVAL retains performance without re-tuning, suggesting that the combination of multimodal cues and constrained outputs provides invariances that standard classifiers lack.
\textbf{(3) Transfer without per-city fine-tuning.} Using a single global model and fixed thresholds, PVAL maintains high weighted averages across all six cities, while baseline spreads increase with distance from the training domain. This indicates the method’s practical deployability in new countries where labeled data and computing are scarce.

Across the global set, a practical residual errors concentrate in dense industrial arrays with specular highlights, small residential arrays near albedo-matched surfaces, and scenes with occlusions from shadows or rooftop equipment. 
}

\section{Applications in Power Systems}
{\color{black}
Rooftop–PV classification yields geolocated, schema-constrained fields (presence, location, quantity). These signals plug naturally into two core distribution-system workflows that are well documented in the literature: 
(i) \emph{PV-aware distribution system state estimation (DSSE)} and voltage management, where restoring behind-the-meter (BTM) injections improves observability and accuracy \cite{Qing2025DSSEReview,Zhi2024DynDSSE}; and 
(ii) \emph{short-term net-load forecasting}, where modeling or separating BTM PV is known to reduce mid-day errors and ramp mispredictions \cite{Stratman2023NetLoadBTM,Qu2025UnsupervisedDisagg,INL2024BTMMonitoring}. 
Guided by these precedents, we instantiate two case studies that mirror microgrid/ADN practice: a PV-aware DSSE example and a PV-aware one-step net-load forecaster. In both, PVAL’s outputs are used exactly as operational features (PV presence indicators and simple PV proxies), illustrating an end-to-end path from imagery to actionable grid signals.
}

{\color{black}
\subsection{State Estimation}
\label{sec:se}
\begin{figure}[htbp]
    \centering
    \color{black}
    \includegraphics[width=1\linewidth]{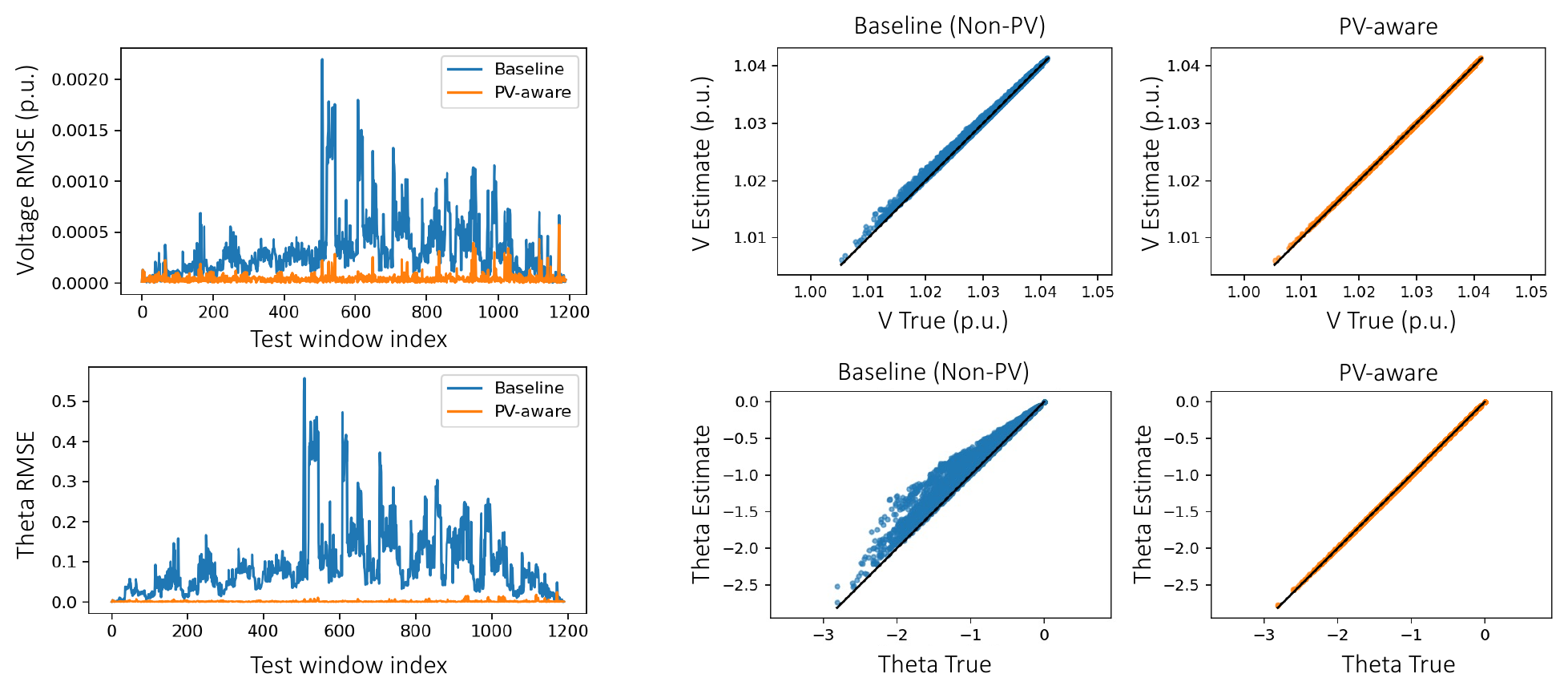}
    \caption{\textbf{PV-aware state estimation reduces bias and error.} Left: PV-aware RMSE for $V$ and $\theta$ remains low across test windows (1200 samples); Non-PV exhibits mid-day spikes. Right: estimates vs.\ truth concentrate along $y{=}x$ for PV-aware, indicating near-unbiased states.}
    \label{fig:StateEstimation}
\end{figure}

In real feeders, meters observe \emph{net} injections $P^{\text{net}}{=}P^{\ell}-P^{\text{pv}}$ and $Q^{\text{net}}{=}Q^{\ell}-Q^{\text{pv}}$. If SE treats $P^{\text{net}}$ as pure load, the injection vector is biased; under LinDistFlow, $\Delta V \!\approx\! H_P\Delta P + H_Q\Delta Q$, so substituting $\Delta P^{\text{net}}$ introduces the error $H_P(-\Delta P^{\text{pv}})$ and systematically biases voltages and angles. Restoring PV as an explicit injection, $(P^{\text{net}}{+}P^{\text{pv}},Q)$, removes this bias.

On an 8-bus feeder, we train a fast surrogate $(P,Q)\!\mapsto\!(V,\theta)$ on 60\% of the data and test on 40\%. Hidden rooftop PV is synthesized on buses $\{3,5,6,7\}$ with a diurnal half-sine (40\% of local load at peak). We compare a \emph{Non-PV} estimator that uses $(P^{\text{net}},Q)$ with a \emph{PV-aware} one that adds PV back before estimation. Figure~\ref{fig:StateEstimation} shows uniformly lower and flatter RMSE for $V$ and $\theta$ and tighter alignment with the identity line, while Non-PV errors spike at mid-day when netting hides injections. 

PV-aware state estimation improves feeder planning (how much new PV the feeder can host and whether upgrades are needed), makes day-to-day voltage control more stable (fewer unnecessary changes to transformer tap settings and capacitor banks), enables clearer city-scale mapping when combined with rooftop-PV detection (native demand, PV, and states in one view), and strengthens policy and billing checks such as net-metering audits and feeder-loss studies.

\subsection{Load Forecasting}
\label{sec:load-forecasting}
\begin{figure}[htbp]
    \centering
    \color{black}
    \includegraphics[width=1\linewidth]{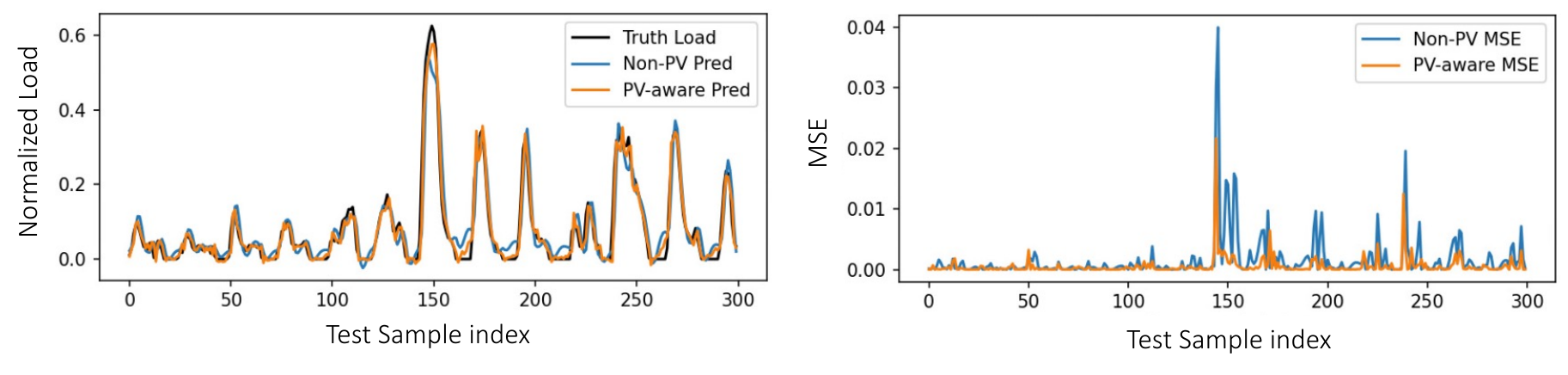}
    \caption{\textbf{PV-aware exogenous signals stabilize one-step net-load forecasts.}
    Left: truth and predictions over the first 300 hourly test samples.
    Right: per-step MSE for the same 300 samples.}
    \label{fig:LoadForecasting}
\end{figure}

Forecasters trained only on net load $y_t^{\text{net}}{=}d_t{-}p_t^{\text{pv}}$ never see how PV changes during the day, so they tend to make large errors around noon and during fast ramps. If we also give the model PV information, those errors should shrink.

We use hourly residential data from Houston~\cite{ref:kaggle_residential}. Where applicable, we add realistic diurnal PV profiles to form a \emph{net} series and forecast one step ahead using a look-back window of $L{=}168$ hours. The \emph{Non-PV LSTM} sees only past net load. The \emph{PV-aware LSTM} sees the same net history plus PV history, a simple one-step PV forecast $\hat p_{t+1}^{\text{pv}}$ (persistence), and a flag indicating whether PV is present. Both models use the same architecture, training procedure, scaling, and chronological train/validation/test splits, and we report errors on the original (de-normalized) scale.

As shown in Figure~\ref{fig:LoadForecasting}, we visualize the first 300 hourly samples from the test split. The PV-aware model follows peaks and afternoon ramps more closely and greatly reduces the mid-day MSE spikes, while night-time accuracy remains similar. For local grids and microgrids, this means tighter reserve planning, better day-ahead bids and demand-response targeting, and more reliable short-term checks on transformer loading when rooftop PV is significant.
}

\section{Limitations and Future Work}
{\color{black}
While the PVAL framework shows strong performance and generalization, several limitations remain. First, as with most LLM-based systems, the model may inherit biases from pre-training data or uneven regional coverage, leading to variations across socioeconomic or geographic contexts. Second, reliance on commercial imagery (e.g., Google Maps) introduces copyright and licensing restrictions; although we share metadata and scripts, users must regenerate imagery with their own API keys. Third, rooftop-level analysis raises privacy concerns, requiring future work on privacy-preserving mechanisms and regulatory compliance. Fourth, the structured JSON outputs improve interpretability but limit the use of standard vision metrics (IoU, mAP). Finally, scalability has not yet been tested on million-scale imagery or deployment-level simulations, which are critical for real-world robustness.

Future work will address these limitations by developing bias detection and mitigation methods, expanding the schema to capture richer PV attributes, integrating pixel-level postprocessing for benchmarking, and conducting large-scale deployment studies to validate scalability and operational utility.
}

\section{Conclusion}
This paper {\color{black}introduces PVAL, a label-efficient framework that unifies task decomposition and schema-guided JSON with \emph{likelihood}/\emph{confidence}, along with few-shot prompting and targeted fine-tuning, to} efficiently detect {\color{black}, localize, and count rooftop} solar panels in satellite imagery{\color{black}, achieving an overall accuracy of 94\% across six U.S. regions and surpassing all baselines}, reducing reliance on extensive labeled datasets and computational resources. 
The proposed PVAL framework surpasses existing methods with superior scalability across diverse environments and adaptability to multiple tasks{\color{black}; cross-continental experiments show robust transfer across urban forms and imaging conditions}. 
{\color{black}Practically, PVAL’s standardized, machine-actionable outputs integrate directly with GIS/EMS to maintain PV inventories, screen interconnection requests, and support feeder planning, hosting-capacity analysis, and DER siting. }
These results demonstrate the potential of {\color{black}constraining multimodal} LLMs {\color{black}with a spatial schema and uncertainty signals as} cost-effective tools to accelerate renewable energy integration and enhance the resilience of microgrids and active ADNs.

% {\color{black}
% We introduced PVAL, a label-efficient framework that unifies task decomposition, schema-guided JSON outputs with likelihood/confidence, few-shot prompting, and targeted fine-tuning to detect, localize, and count rooftop PV from satellite imagery. Cross-continental experiments show robust transfer across urban forms and imaging conditions, while a confidence-driven auto-labeling mechanism reduces annotation effort and enables scalable dataset growth. 
% Practically, PVAL’s standardized, machine-actionable outputs integrate directly with GIS/EMS to maintain PV inventories, screen interconnection requests, and support feeder planning, hosting-capacity analysis, and DER siting. More broadly, constraining multimodal LLMs with a spatial schema and uncertainty signals provides a cost-effective path to deployment-grade PV mapping for microgrids and ADNs, accelerating renewable integration and grid resilience.
% }

\section*{Acknowledgments}
The author expresses sincere gratitude to Yizheng Liao for his insightful suggestions regarding the utilization of prompt engineering and fine-tuning. We also acknowledge meaningful suggestions from Dr. Erik Blasch.
\bibliographystyle{elsarticle-num} 
\bibliography{Bibliography}




% %% For numbered reference style

%   \textit{\LaTeX: a document preparation system},
%   Addison Wesley, Massachusetts,

% \end{thebibliography}

\end{document}